\documentclass{article} 
\usepackage{nips14submit_e,times}
\usepackage{hyperref}
\usepackage{url}

\usepackage{amssymb,amsfonts,amsmath,amsthm,amscd,dsfont,mathrsfs}
\usepackage{graphicx,float,psfrag,epsfig}
\usepackage{wrapfig}
\usepackage{algorithm,algorithmic,hyperref}

\DeclareMathAlphabet{\mathpzc}{OT1}{pzc}{m}{it}

\newtheorem{propo}{Proposition}[section]
\newtheorem{lemma}[propo]{Lemma}

\newtheorem{thm}{Theorem}
\newtheorem{theorem}[propo]{Theorem}
\newtheorem{remark}[propo]{Remark}

\def\cE{{\cal E}}
\def\cV{{\cal V}}

\def\tP{{\tilde P}}
\def\tpi{{\tilde \pi}}

\def\dmax{d_{\rm max}}

\def\<{\langle}
\def\>{\rangle}

\def\prob{\mathbb P}
\def\reals{\mathbb R}
\def\diag{{\rm diag}}

\def\bX{\bar{X}}
\def\cP{{\cal P}}
\def\bq{{\bf q}}
\def\bw{{\bf w}}
\def\hw{{\hat w}}

\def\tSigma{\tilde{\Sigma}}
\def\hSigma{\hat{\Sigma}}
\def\hU{\hat U}
\def\tU{\tilde U}
\def\tH{\tilde H}
\def\hsm{\hSigma_{M_2}}

\def\hv{{\hat v}}
\def\hnu{\hat{\nu}}
\def\hA{\hat{A}}
\def\tp{\tilde p}

\def\reals{{\mathbb R}}
\def\prob{{\mathbb P}}
\def\E{\mathbb E}

\def\id{{\mathds I}}

\def\r{r}
\def\l{\ell}
\def\n{N}
\def\cS{{\cal S}}
\def\M{M}
\def\hM{\hat{M}}
\def\tM{\tilde{M}}
\def\matrixcompletion{{\sc Matrix Completion}}

\def\w{{w}}

\def\hP{\hat{P}} 
\def\P{P}
\def\hq{\hat{q}}
\def\q{q}
\def\qmin{q_{\rm min}}
\def\qmax{q_{\rm max}}
\def\Q{Q}
\def\U{U}
\def\hQ{\hat{Q}}
\def\eps{\varepsilon}
\def\tv{\tilde v}
\def\bv{\bar v}
\def\V{V}
\def\hV{\hat V}
\def\tV{\widetilde V}
\def\bV{\bar V}
\def\G{H}
\def\tG{\widetilde G}
\def\hG{\tilde \G}
\def\bG{\bar G}
\def\hlambda{\hat \lambda}
\def\hLambda{\hat{\Lambda}}

\def\hH{\hat H}
\def\cE{{\cal E}}

\def\matrixcompletion{{\text{\sc{MatrixAltMin}}}}
\def\tensorLS{{\text{\sc TensorLS}}}
\def\dmax{d_{\rm max}}
\def\dmin{d_{\rm min}}
\def\b{b}
\def\bd{\bar d}

\nipsfinalcopy 

\title{Learning Mixed Multinomial Logit Model from Ordinal Data}

\author{
Sewoong Oh\\
Dept. of Industrial and Enterprise Systems Engr.\\
University of Illinois at Urbana-Champaign\\
Urbana, IL 61801\\
\texttt{swoh@illinois.edu}
\And
Devavrat Shah\\
Department of Electrical Engineering\\
Massachussetts Institute of Technology\\
Cambridge, MA 02139\\
\texttt{devavrat@mit.edu}
}
\begin{document}
\maketitle

\begin{abstract}

Motivated by generating personalized recommendations using ordinal (or preference) data, we study the 
question of learning a mixture of MultiNomial Logit (MNL) model, a parameterized class of distributions over
permutations, from partial ordinal or preference data (e.g. pair-wise comparisons). Despite its long standing 
importance across disciplines including social choice, operations research and revenue management, 
little is known about this question. 
In case of single MNL models (no mixture), computationally and statistically 
tractable learning from pair-wise comparisons is feasible.  
However, even learning mixture with two MNL components is infeasible in general.

Given this state of affairs, we seek conditions under which it is feasible to learn the mixture model in both
computationally and statistically efficient manner. We present a sufficient condition as well as 
an efficient algorithm for learning mixed MNL models from partial preferences/comparisons data. In particular, 
a mixture of $r$ MNL components over $n$ objects can be learnt using samples whose size scales polynomially in 
$n$ and $r$ (concretely, $r^{3.5} n^3 (\log n)^4$, with $r \ll n^{2/7}$ when 
the model parameters are sufficiently {\em incoherent}). The algorithm has two phases: first, learn the pair-wise marginals for each
component using tensor decomposition; 
second, learn the model parameters 
for each component using {\sc RankCentrality} introduced by Negahban et al.
In the process of proving these results, 
we obtain a generalization of existing analysis for tensor decomposition 
to a more realistic regime where only partial information about each sample is available. 

\end{abstract}

%
%
\section{Introduction}
\label{sec:intro}

\noindent{\bf Background.} Popular recommendation systems such as collaborative filtering are based on 
a partially observed {\em ratings} matrix. The underlying hypothesis is that the true/latent 
score matrix is low-rank and we observe its partial, noisy version. Therefore, matrix completion algorithms
are used for learning, cf. \cite{CR09,KMO10,kmo10noise,NW11}. 
In reality, however, observed preference data is not just scores. For example, clicking one of the many 
choices while browsing provides partial order between clicked choice versus other choices. Further, 
scores do convey ordinal information as well, e.g. score of 4 for paper A and score of 7 for paper B by a reviewer 
suggests ordering B $> $ A. Similar motivations led Samuelson to propose the {\em Axiom
of revealed preference} \cite{Samuelson38} as the model for rational behavior. In a nutshell, it states that 
{\em consumers have latent order of all objects, and the revealed preferences through actions/choices 
are consistent with this order.} If indeed all consumers had identical ordering, then learning preference
from partial preferences is effectively the question of sorting. 

In practice, individuals have different orderings of interest, and further, each individual is likely to make noisy 
choices. 
This naturally suggests the following model -- each individual has a latent distribution over orderings 
of objects of interest, and the revealed partial preferences are consistent with it, i.e. samples from the
distribution. Subsequently, the preference of the population as a whole can be associated with a distribution over 
permutations. 
Recall that the low-rank structure for score matrices, as a model, tries to capture the fact that there are only a few different
types of choice profile. In the context of modeling consumer choices as distribution over permutation, 
MultiNomial Logit (MNL) model with a small number of  mixture components provides such a model. 

\noindent{\bf Mixed MNL.}  Given $n$ objects or choices of interest, an MNL model is described as a parametric 
distribution over permutations of $n$ with  parameters $\bw = [w_i]\in\reals^n$: each object $i, ~1\leq i\leq n$, has a parameter $w_i > 0$
associated with it. Then the permutations are generated randomly as follows: choose one of the $n$ objects to be
ranked $1$ at random, where object $i$ is chosen to be ranked $1$ with probability $w_i/(\sum_{j=1}^n w_j)$. Let $i_1$
be object chosen for the first position. Now to select second ranked object, choose from remaining with probability
proportional to their weight. We repeat until all objects for all ranked positions are chosen. It can be easily seen that, as per
this model, an item $i$ is ranked higher than $j$ with probability $w_i/(w_i + w_j)$. 

In the mixed MNL model with $r \geq 2$ mixture components, each component corresponds to a different 
MNL model: let $\bw^{(1)},\dots, \bw^{(r)}$ be the corresponding parameters of the $r$ components. Let
$\bq = [q_a] \in [0,1]^r$ denote the mixture distribution, i.e. $\sum_a q_a = 1$. To generate
a permutation at random, first choose a component $a \in \{1,\dots, r\}$ with probability $q_a$, 
and then draw random permutation as per MNL with parameters $\bw^{(a)}$.

\noindent{\bf Brief history.} The MNL model is an instance of a class of models introduced by
Thurstone \cite{Thu27}. The description of the MNL provided here was formally established
by McFadden  \cite{McF73}. The same model (in form of pair-wise marginals) was introduced by  Zermelo \cite{Zer29} as well as 
Bradley and Terry \cite{BT55} independently. In \cite{Luce59}, Luce established that MNL is the only distribution
over permutation that satisfies the axiom of Independence from Irrelevant Alternatives. 

On learning distributions over permutations,  
the question of learning single MNL model and
more generally instances of Thurstone's model have been of interest for quite a while now. 
The maximum likelihood estimator, which is logistic regression for MNL, has been known to be consistent in large sample
limit, cf. \cite{Ford57}. Recently, {\sc RankCentrality} \cite{NOS12} was established to be statistical efficient. For
learning sparse mixture model, i.e. distribution over permutations with each mixture being delta distribution, 
\cite{FJS09} provided sufficient conditions under which mixtures can be learnt exactly using pair-wise marginals -- 
effectively,  as long as the number of components scaled as $o(\log n)$ where components satisfied appropriate
{\em incoherence} condition, a simple iterative algorithm could recover the mixture. However, it is not robust
with respect to noise in data or finite sample error in marginal estimation. Other approaches have been
proposed to recover model using convex optimization based techniques, cf. \cite{DMJ10, MGCV11}.
MNL model is a special case of a larger family of discrete choice models known as the Random Utility Model (RUM), 
and an efficient algorithm to learn RUM  is introduced in \cite{SPX12}. 
Efficient algorithms for learning RUMs from 
partial rankings has been introduced in \cite{ACPX13,APX14}. 
We note that the above list of references is very limited, including only closely related literature. Given the nature of the topic, there
are a lot of exciting lines of research done over the past century and we shall not be able to provide comprehensive coverage
due to a space limitation.

\noindent{\bf Problem.} Given observations from the mixed MNL, we wish to learn the model parameters, 
the mixing distribution $\bq$, and parameters of each component $\bw^{(1)}, \dots, \bw^{(r)}$. The observations
are in form of pair-wise comparisons. Formally, to generate an observation, first one of the $r$ 
mixture components is chosen; and then for $\ell$ of all possible ${n \choose 2}$ pairs, comparison 
outcome is observed as per this MNL 
component\footnote{We shall assume that, outcomes of  these $\ell$ pairs are independent of
each other, but coming from the same MNL mixture component. This is effectively true even they were generated by first sampling a permutation 
from the chosen MNL mixture component, and then observing implication of this permutation for the specific $\ell$ pairs, as long as they are
distinct due to the Irrelevance of Independent Alternative hypothesis of Luce that is satisfied by MNL.}. 
These $\ell$ pairs are chosen, uniformly at random, from a pre-determined 
$N \leq {n \choose 2}$ pairs: $\{(i_k, j_k), 1\leq k \leq N\}$. We shall assume that the selection of $N$ is such that 
the undirected graph $G=([n], E)$, where $E = \{(i_k, j_k): 1\leq k\leq N\}$, is connected.   

We ask following questions of interest: Is it always feasible to learn mixed MNL? If not, under what conditions and how many samples
are needed? How computationally expensive are the algorithms? 

\begin{figure}[h!]\label{fig1}
	\begin{center}
	\vspace{-.7cm}
	\includegraphics[width=.5\textwidth]{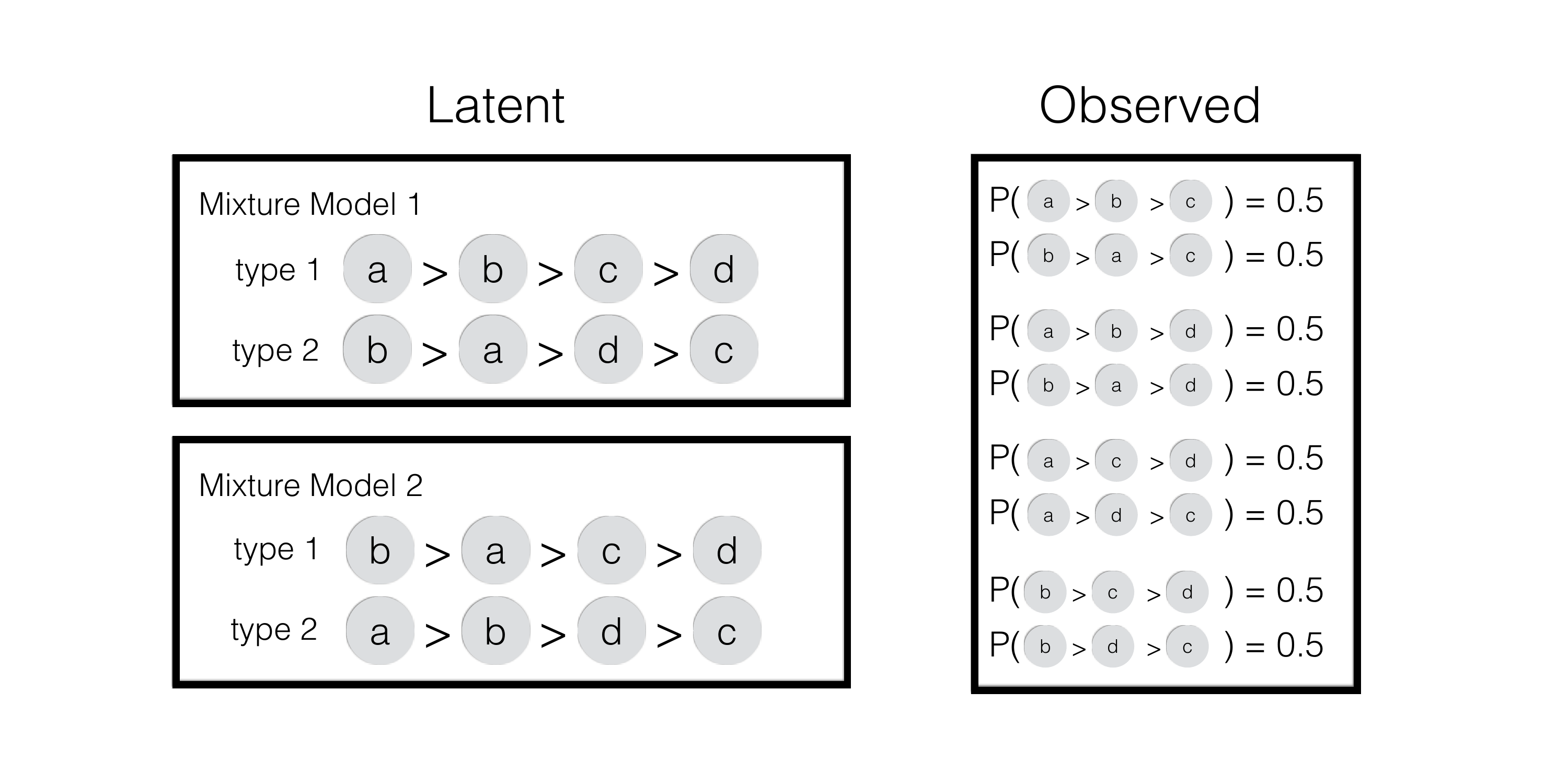}
	\vspace{-.7cm}
	\caption{Two mixture models that cannot be differentiated even with $3$-wise preference data.}
	\end{center}
\end{figure}

We briefly recall a recent result  \cite{AOSV14} that suggests that it is impossible to learn mixed MNL models in general. 
One such example is described in Figure \ref{fig1}. It depicts an example with $n = 4$ and $r=2$ and a uniform mixture distribution. 
For the first case, in mixture component $1$, with probability $1$ the ordering is $a > b > c > d$ (we denote
$n =4$ objects by $a, b, c$ and $d$); and in mixture component $2$, with probability $1$ the ordering is 
$b > a > d > c$. Similarly for the second case, the two mixtures are made up of permutations $b > a > c > d$ and 
$a > b > d > c$. It is easy to see the distribution over any $3$-wise comparisons generated from these two mixture
models is identical. Therefore, it is impossible to differentiate these two using $3$-wise or pair-wise comparisons. 
In general, \cite{AOSV14} established that there exist mixture distributions with $r \leq n/2$ over $n$ objects that
are impossible to distinguish using $\log n$-wise comparison data. That is, learning mixed MNL is not 
always possible. 

\vspace{-0.5cm}
\noindent{\bf Contributions.} The main contribution of this work is identification of sufficient conditions under which 
mixed MNL model can be learnt efficiently, both statistically and computationally. Concretely, we propose a two-phase
learning algorithm: in the first phase, using a tensor decomposition method for learning mixture of discrete product
distribution, we identify pair-wise marginals associated with each of the mixture; in the second phase, we use these pair-wise 
marginals associated with each mixture to learn the parameters associated with each of the MNL mixture component. 

The algorithm in the first phase builds upon the recent work by Jain and Oh \cite{JO14}. In particular, Theorem \ref{thm:finite}
generalizes their work for the setting where for each sample, we have limited information - as per \cite{JO14}, we would
require that each individual gives the entire permutation; instead, we have extended the result to be able to cope with the current setting when we
only have information about $\ell$, potentially finite, pair-wise comparisons. The algorithm in the second phase utilizes
{\sc RankCentrality} \cite{NOS12}. Its analysis in Theorem \ref{thm:rankcentrality} works for setting where observations are
no longer independent, as required in \cite{NOS12}. 

We find that as long as certain rank and {\em incoherence} conditions are satisfied by the parameters of each of the mixture, 
the above described two phase algorithm is able to learn mixture distribution $\bq$ and parameters associated with
each mixture, $\bw^{(1)}, \dots, \bw^{(r)}$ faithfully using samples that scale polynomially in $n$ and $r$ -- concretely, the
number of samples required scale as $r^{3.5} n^3 (\log n)^4$ with constants dependent on the incoherence between
mixture components, and as long as $r \ll n^{2/7}$ as well as $G$, the graph of potential comparisons, is a spectral 
expander with the total number of edges scaling as $\n = O(n \log n)$. For the precise statement, we refer to 
Theorem \ref{thm:main}.

The algorithms proposed are iterative, and primarily based on spectral properties of underlying tensors/matrices with
provable, fast convergence guarantees. That is, algorithms are not only polynomial time, they are practical enough
to be scalable for high dimensional data sets.


\noindent{\bf Notations.} We use $[N]=\{1,\ldots,N\}$ for the first $N$ positive integers. 
We use  $\otimes$ to denote the outer product such that $(x\otimes y\otimes z)_{ijk}= x_iy_jz_k$.
Given a third order tensor $T\in\reals^{n_1\times n_2\times n_3}$ 
and a matrix $U\in\reals^{n_1\times r_1},V\in\reals^{n_2\times r_2},W\in\reals^{n_3\times r_3}$, 
we define a linear mapping $T[U,V,W]\in\reals^{r_1\times r_2\times r_3}$ as $T[U,V,W]_{abc}=\sum_{i,j,k} T_{ijk}U_{ia}V_{jb}W_{kc} $. 
We let  $\| x \|=\sqrt{\sum_i x_i^2}$ be the Euclidean norm of a vector, 
 $\|M\|_2=\max_{\|x\|\leq 1,\|y\|\leq 1}x^TMy$ be the operator norm of a matrix, 
 and $\|M\|_F=\sqrt{\sum_{i,j} M_{ij}^2}$ be the Frobenius norm.  
We say an event happens with high probability (w.h.p) if the probability is lower bounded by $1-f(n)$ such that $f(n)=o(1)$ as $n$ 
scales to $\infty$.

\section{Main result}

In this section, we describe the main result: sufficient conditions under which mixed MNL
models can be learnt using tractable algorithms. We provide a useful illustration of the result as well as
discuss its implications.  

\noindent {\bf Definitions.} Let $\cS$ denote the collection of observations, each of which is
denoted as $N$ dimensional, $\{-1, 0, +1\}$ valued vector. Recall that each observation is obtained by 
first selecting one of the $r$ mixture MNL component, and then viewing outcomes, as per the chosen
MNL mixture component, of $\ell$ randomly chosen pair-wise comparisons from the $N$ pre-determined 
comparisons $\{(i_k, j_k): 1\leq i_k\neq j_k \leq n,  1\leq k\leq N\}$. Let $x_t \in \{-1, 0, +1\}^N$ denote  the $t$th 
observation with $x_{t,k} = 0$ if the $k$th pair $(i_k,j_k)$ is not chosen amongst the $\ell$ randomly chosen pairs, 
and $x_{t,k} = +1$ (respectively $-1$) if $i_k < j_k$ (respectively $i_k > j_k$) as per the chosen MNL mixture component. 
By definition, it is easy to see that for any $t \in \cS$ and $1\leq k\leq N$, 
\begin{align}\label{eq:defP}
\E[x_{t,k}] & = \frac{\ell}{N} \Big[\sum_{a=1}^r q_a P_{ka}\Big], ~\mbox{where}~P_{ka} = \frac{w_{j_k}^{(a)} - w_{i_k}^{(a)}}{w_{j_k}^{(a)} + w_{i_k}^{(a)}}.
\end{align}
We shall denote $P_a = [P_{ka}] \in [-1,1]^N$ for $1\leq a\leq r$. Therefore, in a vector form 
\begin{align}\label{eq:1}
\E[x_{t}] & = \frac{\ell}{N}P\bq, ~\mbox{where}~~P = [P_1 \dots P_r] \in [-1,1]^{N\times r}\;. 
\end{align}
That is, $P$ is a matrix with $r$ columns, each representing one of the $r$ mixture components and 
$\bq$ is the mixture probability.
By independence, for any $t \in \cS$, and any  two different pairs $1\leq k \neq m \leq N$,
\begin{align}
\E[x_{t,k} x_{t,m}] & =\frac{\ell^2}{N^2} \Big[\sum_{a=1}^r q_a P_{ka}P_{ma} \Big].
\end{align}
Therefore, the $N\times N$ matrix $\E[x_t x_t^T]$ or equivalently tensor $\E[x_t \otimes x_t]$ 
is proportional to $M_2$ except in diagonal entries, where 
\begin{align}\label{eq:2}
M_2 & = PQP^T \equiv \sum_{a=1}^r q_a (P_a \otimes P_a)\;,
\end{align}
$Q = \diag(\bq)$ being diagonal matrix with its entries being mixture probabilities, $\bq$. 
In a similar manner, the tensor $\E[x_t \otimes x_t \otimes x_t]$ is proportional to $M_3$ (except in $O(N^2)$ 
entries), where 
\begin{align}\label{eq:3}
M_3 & = \sum_{a=1}^r q_a (P_a \otimes P_a \otimes P_a). 
\end{align}
Indeed, empirical estimates $\hat{M}_2$ and $\hat{M}_3$, defined as 
\begin{align}\label{eq:4}
\hat{M}_2 & = \frac{1}{|\cS|} \Big[\sum_{t\in \cS} x_t \otimes x_t\Big], ~\mbox{and}~\hat{M}_3  = \frac{1}{|\cS|} \Big[\sum_{t\in \cS} x_t \otimes x_t \otimes x_t\Big],
\end{align} 
provide good proxy for $M_2$ and $M_3$ for large enough number of samples; and shall be utilized crucially for learning model parameters from observations. 

\noindent {\bf Sufficient conditions for learning.} With the above discussion, we state sufficient conditions for learning the mixed MNL in terms of properties of $M_2$: 
\begin{itemize}
\item[C1.]  $M_2$ has rank $r$; let $\sigma_1(\M_2)$, $\sigma_r(\M_2)>0$ be the largest and smallest singular values of $M_2$. 
\item[C2.]  For a large enough universal constant $C' > 0$, 
\begin{align} 
N &\geq C' {\r^{3.5}\,\mu^6(M_2)}\,\Big(\frac{\sigma_1(\M_2)}{\sigma_\r(\M_2)}\Big)^{4.5}.
\end{align}
In the above, $\mu(M_2)$ represents incoherence of a symmetric matrix $M_2$. We recall that for
a symmetric matrix $M\in\reals^{\n\times\n}$ of rank $\r$ with singular value decomposition 
$M=USU^T$, the incoherence is defined as 
\begin{align}\label{eq:defmu}
	\mu(M) &= \sqrt{\frac{\n}{r}} \Big(\max_{i\in[\n]} \|U_i\|\Big).
\end{align}
\item[C3.] The undirected graph $G = ([n], E)$ with $E = \{(i_k, j_k): 1\leq k\leq N\}$ is 
connected. Let $A \in \{0,1\}^{n \times n}$
be adjacency matrix with $A_{ij} = 1$ if $(i,j) \in E$ and $0$ otherwise; let $D = \diag(d_i)$ with $d_i$ being degree of
vertex $i \in [n]$ and let $L_G = D^{-1} A$ be normalized Laplacian of $G$. 
Let $\dmax = \max_i d_i$ and $\dmin = \min_i d_i$. 
Let the $n$ eigenvalues of 
stochastic matrix $L_G$ be $1 = \lambda_1(L_G) \geq \dots \lambda_n(L_G) \geq -1$. Define spectral gap of 
$G$: 
\begin{align}\label{eq:defxi}
\xi(G) & = 1 - \max\{\lambda_2(L),-\lambda_n(L)\}.
\end{align}
\end{itemize}
Note that we choose a graph $G=([n],E)$ to collect pairwise data on, 
and we want to use a graph that is connected, has a large spectral gap, and has a small number of edges. 
In condition (C3),  
we need connectivity  since we cannot estimate the relative strength between disconnected components (e.g. see \cite{Ford57}).
Further, it is easy to generate a graph with spectral gap $\xi(G)$ bounded below by a universal constant (e.g. $1/100$) and the number of edges $\n=O(n\log n)$, 
for example using the configuration model for Erd\"os-Renyi graphs. 
In condition (C2), we require the matrix $\M_2$ to be sufficiently incoherent with bounded $\sigma_1(\M_2)/\sigma_\r(\M_2)$.
For example, if $\qmax/\qmin = O(1)$ and the profile of each type in the mixture distribution is sufficiently 
different, i.e. $\<P_a,P_b\>/(\|P_a\|\|P_b\|)<1/(2r)$, then we have $\mu(M_2)=O(1)$ 
and $\sigma_1(M_2)/\sigma_r(M_2) = O(1)$. 
We define $b = \max_{a=1}^r \max_{i,j \in [n]} {w^{(a)}_i}/{w^{(a)}_j}$, $\qmax = \max_a q_a$, and  $\qmin = \min_a q_a$.  
The following theorem provides a  bound on the error and we refer to the appendix for a proof. 

\begin{thm}
\label{thm:main}
	Consider a mixed MNL model satisfying conditions (C1)-(C3). Then for any $\delta\in(0,1)$, 
	there exists positive numerical constants $C,C'$ such that for any positive $\varepsilon$ satisfying  
	\begin{align} 
	0 < \varepsilon <~ \Big(\frac{\qmin \xi^2(G) \dmin^2}{16 \qmax\,\r\,\sigma_1(\M_2) b^5 \dmax^2}\Big)^{0.5},
	\end{align}
	Algorithm \ref{algo:main} produces estimates $\hat{\bq} = [\hat{q}_a]$ and $\hat{\bw} = [\hat{\bw}^{(a)}]$ so that with probability at least $1-\delta$, 
	\begin{align}
		\big| \;\hq_a -q_a\;\big| &\leq \varepsilon,  \text{ and }\nonumber \\
		\frac{\| \hat{\bw}^{(a)} -\bw^{(a)}\|}{\|\bw^{(a)}\|} &\leq C\Big(\frac{r\,\qmax\,\sigma_1(\M_2) \b^{5} \dmax^2 }{\qmin \xi^2(G) \dmin^2}\Big)^{0.5} ~ \varepsilon, 
	\end{align}
	for all $a\in[\r]$, as long as 
	\begin{align}
	|\cS| &\geq C' \frac{r \n^4 \log(\n/\delta)}{\qmin \sigma_1(\M_2)^2\varepsilon^2} \Big(\frac{1}{\l^2} +\frac{\sigma_1(\M_2)}{\l \n}+\frac{r^4\sigma_1(\M_2)^4}{\sigma_r(\M_2)^5}\Big)\;.
	\end{align}	
\end{thm}

\noindent{\bf An illustration of  Theorem \ref{thm:main}.} 
To understand the applicability of Theorem \ref{thm:main}, consider a concrete example
with $r =2$; let the corresponding weights $\bw^{(1)}$ and $\bw^{(2)}$ be generated by choosing each weight uniformly
from $[1,2]$. In particular, the rank order for each component is a uniformly random permutation. 
Let the mixture distribution be
uniform as well, i.e. $\bq = [0.5~ 0.5]$. Finally, let the graph $G = ([n], E)$ be chosen as per the Erd\"os-R\'enyi model with
each edge chosen to be part of the graph with probability $\bd/n$, where $\bd > \log n$.  
For this example, it can be checked that Theorem \ref{thm:main} guarantees that for $\varepsilon \leq C/\sqrt{n \bd}$, 
$|\cS|\geq C' n^2\bd^2\log(n\bd/\delta)/(\l\varepsilon^2)$,  and $n \bd \geq C'$, we have for all $a\in\{1,2\}$, 
$|\hq_a- q_a|\leq\varepsilon$ and $\|\hat{\bw}^{(a)}-\bw^{(a)}\|/\|\bw^{(a)}\| \leq C''\sqrt{n \bd}\,\varepsilon$. 
That is, for $\ell =\Theta(1)$ and choosing $\varepsilon = \varepsilon'/(\sqrt{n \bar{d}})$, we need sample size 
of $|\cS| = O(n^3\bd^3 \log n)$ to guarantee error in both $\hat{\bq}$ and $\hat{\bw}$ smaller than $\varepsilon'$. 
Instead, if we choose $\ell = \Theta(n \bar{d})$, we only need  $|\cS| = O((n\bd)^2 \log n) $. 
Limited samples per observation leads to penalty of factor of $(n\bd/\l)$ in sample complexity. 
To provide bounds on the problem parameters for this example,  we  use standard concentration arguments. 
It is well known for Erd\"os-R\'enyi random graphs (see \cite{Bol01}) that, with high probability, the number of edges 
concentrates in  $[(1/2)\bd\,n, (3/2)\bd\, n]$ implying $\n=\Theta(\bd n)$, and the degrees also concentrate in $[(1/2)\bd,(3/2)\bd]$, 
implying $\dmax=\dmin=\Theta(\bd)$. Also using standard concentration arguments for spectrum of random matrices, 
it follows that the spectral gap of $G$ is bounded by $\xi\geq 1-(C/\sqrt{\bd})=\Theta(1)$ w.h.p. Since we assume the weights to 
be in $[1,2]$, the dynamic range is bounded by $\b\leq 2$.  The following Proposition shows that $\sigma_1(\M_2)=\Theta(\n)=\Theta(\bd n)$,
$\sigma_2(\M_2)=\Theta(\bd n)$, and $\mu(\M_2)=\Theta(1)$. 
\begin{propo}\label{rem:example}
	For the above example, when $\bd \geq \log n$, $\sigma_1(\M_2)\leq 0.02 \n$, $\sigma_2(\M_2)\geq 0.017 \n$, 
	and $\mu(\M_2)\leq 15$ with high probability. 
\end{propo}
Supposen now for general $r$, we are interested in well-behaved scenario where 
$\qmax=\Theta(1/r)$ and $\qmin=\Theta(1/\r)$.
To achieve arbitrary small error rate for  $\| \hat{\bw}^{(a)} -\bw^{(a)}\|/\|\bw^{(a)}\|$, 
we need $\epsilon=O(1/\sqrt{r\,N})$, which is achieved by 
sample size $|\cS|=O( r^{3.5} n^3 (\log n)^4)$ with $\bd=\log n$. 

\section{Algorithm}

We describe the algorithm achieving the bound in Theorem \ref{thm:main}. 
Our approach is two-phased. 
First, learn the moments for mixtures using a tensor decomposition, cf. Algorithm \ref{algo:spect}: for each type $a\in[\r]$, produce 
estimate $\hq_a\in\reals$ of the mixture weight $q_a$ and estimate $\hP_a=[\hP_{1a} \,\ldots\,\hP_{\n a}]^T\in\reals^{\n}$ of 
the expected outcome $\P_a=[\P_{1a} \, \ldots \, \P_{\n a}]^T$ defined as in \eqref{eq:defP}. 
Secondly, for each $a$, using the estimate $\hP_a$,  apply {\sc RankCentrality}, cf. Section \ref{algo:rank}, 
to estimate $\hat{\bw}^{(a)}$ for the MNL weights $\bw^{(a)}$. 
 
\begin{algorithm}[h!]
\caption{}
\label{algo:main}
\begin{algorithmic}[1]
  \STATE {\bf Input}: Samples $\{x_t\}_{t\in\cS}$, number of types $\r$, number of iterations  $T_1,T_2$, graph $G([n],E)$
\STATE $\{(\hq_a,\hP_a)\}_{a\in[\r]}\leftarrow \text{\sc SpectralDist} \left( \{x_t\}_{t\in\cS},\r,T_1 \right)$ \hfill(see Algorithm~\ref{algo:spect})
\FOR{$a=1,\ldots,\r$} 
\STATE set $\tP_a\leftarrow \cP_{[-1,1]}(\hP_a) $ where $\cP_{[-1,1]}(\cdot)$ is the projection onto $[-1,1]^\n$
\STATE $\hw^{(a)}\leftarrow \text{\sc RankCentrality} \left(G,\tP_{a},T_2\right)$ \hfill (see Section \ref{algo:rank})
 \ENDFOR
\STATE {\bf Output}: $\{(\hq^{(a)},\hat{\bw}^{(a)})\}_{a\in[\r]}$ 
\end{algorithmic}
\end{algorithm}
To achieve Theorem \ref{thm:main}, $T_1 = \Theta\big(\log(\n\,|\cS|)\big)$ and 
$T_2=\Theta\big(\,\b^2 \dmax(\log n + \log(1/\varepsilon))/(\xi\dmin)\,  \big)$ is sufficient. 
Next, we describe
the two phases of algorithms and associated technical results.

%
%

\subsection{Phase 1: Spectral decomposition.} 

To estimate $P$ and $\bq$ from the samples, we shall use tensor decomposition of $\hat{M}_2$ and $\hat{M_3}$, 
the empirical estimation of $M_2$ and $M_3$ respectively, recall \eqref{eq:2}-\eqref{eq:4}.
%
%
%
Let $\M_2=U_{\M_2}\Sigma_{\M_2}U_{\M_2}^T$ be the eigenvalue decomposition and  let
\begin{eqnarray*}
\G &=& M_3[U_{\M_2}\Sigma_{\M_2}^{-1/2},U_{\M_2}\Sigma_{\M_2}^{-1/2},U_{\M_2}\Sigma_{\M_2}^{-1/2}]\;.
\end{eqnarray*}
The next theorem shows that $M_2$ and $M_3$ are sufficient to learn $P$ and $\bq$ exactly, when $M_2$ has rank $\r$ (throughout,
we assume that $\r \ll n \leq N$).  
\begin{thm}[Theorem 3.1\,\cite{JO14}]
\label{thm:consistency} 
Let $M_2 \in \reals^{N\times N}$ have rank $\r$. Then there exists an 
orthogonal matrix $V^{\G}=[v^{\G}_1 \; v^{\G}_2 \; \ldots \;v^{\G}_\r]\in\reals^{\r\times\r}$ 
and eigenvalues $\lambda^{\G}_a,~1\leq a \leq r$, such that the orthogonal tensor decomposition of $\G$ is 
\begin{eqnarray*}
	\G &=& \sum_{a=1}^\r \lambda^{\G}_a (v^{\G}_a\otimes v^{\G}_a \otimes v^{\G}_a).
\end{eqnarray*}
Let $\Lambda^{\G}=\diag(\lambda^{\G}_1,\ldots,\lambda^{\G}_\r)$. Then the parameters of the mixture distribution are
\begin{eqnarray*}
	\P \;=\; U_{\M_2}\Sigma^{1/2}_{\M_2} V^{\G} \Lambda^{\G} \;\text{ and }\;\;\;\; 
	\Q \;=\; (\Lambda^{\G} )^{-2}\;.
\end{eqnarray*}
\end{thm}

The main challenge in estimating $\M_2$ (resp. $M_3$) from empirical data are the diagonal entires.  In \cite{JO14}, 
alternating minimization approach is used for matrix completion to find the missing diagonal entries of $\M_2$, 
and used a least squares method for estimating the tensor $\G$ directly from the samples. 
Let $\Omega_2$ denote the set of off-diagonal indices for an $\n\times\n$ matrix
and $\Omega_3$ denote the off-diagonal entries of an $\n\times\n\times\n$ tensor such that the corresponding
projections are defined as 
\begin{eqnarray*}
	\cP_{\Omega_2}(M)_{ij} \;\equiv\; \left\{ \begin{array}{rl} M_{ij}&\text{ if $i\neq j$}\;, \\ 0&\text{ otherwise}\;.\end{array}\right.
	\;\text{ and }\;\;\;\;
	\cP_{\Omega_3}(T)_{ijk} \;\equiv\; \left\{ \begin{array}{rl} T_{ijk}&\text{ if $i\neq j$, $j\neq k$, $k\neq i$}\;, \\ 0&\text{ otherwise}\;.\end{array}\right.
\end{eqnarray*}
for $M \in \reals^{N\times N}$ and $T\in\reals^{\n\times \n\times \n} $.
\begin{algorithm}[thb]
\caption{{\bf {\sc SpectralDist}:} Moment method for Mixture of Discrete Distribution \cite{JO14}}
\label{algo:spect}
\begin{algorithmic}[1]
  \STATE {\bf Input}: Samples $\{x_t\}_{t\in\cS}$, number of types $\r$, number of iterations  $T$
\STATE $\tM_2\leftarrow \matrixcompletion\left(\hat{M}_2\big(1,\frac{|\cS|}{2}\big), r, T \right)$ \hfill(see Algorithm~\ref{algo:altmin})
\STATE Compute eigenvalue decomposition of $\tM_2=\tU_{M_2} \tSigma_{M_2} \tU_{M_2}^T$
\STATE $\hG\leftarrow \tensorLS\left( \hat{M}_3\big(\frac{|\cS|}{2}+1, |\cS|\big), \tU_{M_2}, \tSigma_{M_2} \right)$ \hfill (see Algorithm~\ref{algo:ls})
\STATE Compute rank-$r$  decomposition $\sum_{a\in [\r]} \hlambda_{a}^{\hG} (\hv^{\hG}_a\otimes \hv^{\hG}_a \otimes \hv^{\hG}_a)$ of $\hG$, using RTPM of \cite{AGH12}
\STATE {\bf Output}: $\hP=\tU_{M_2} \tSigma_{M_2}^{1/2} \hV^{\hG} \hLambda^{\hG}$, $\hQ=(\hLambda^{\hG})^{-2}$, where $\hV^{\hG} = [\hv^{\hG}_1\ \dots \ \hv^{\hG}_r]$ and $\hLambda^{\hG} = \diag(\lambda^{\hG}_1,\ldots,\lambda^{\hG}_\r)$
\end{algorithmic}
\end{algorithm}

In lieu of above discussion, we shall use $\cP_{\Omega_2}\big(\hat{M_2}\big)$ and $\cP_{\Omega_3}\big(\hat{M_3}\big)$ to obtain estimation of 
diagonal entries of $M_2$ and $M_3$ respectively. To keep technical arguments simple, we shall use first $|\cS|/2$ samples 
based $\hat{M}_2$, denoted as $\hat{M}_2\big(1,\frac{|\cS|}{2}\big)$ and second $|\cS|/2$ samples based $\hat{M}_3$, denoted
by $\hat{M}_3\big(\frac{|\cS|}{2}+1, |\cS|\big)$ in Algorithm \ref{algo:spect}.

Next, we state correctness of Algorithm \ref{algo:spect} when $\mu(\M_2)$ is small; proof is in Appendix.
\begin{thm}
	\label{thm:finite}
	There exists universal, strictly positive  constants $C,C'>0$ such that 
	for all $\eps\in(0,C)$ and $\delta\in(0,1)$, if 
	\begin{eqnarray*}
	|\cS| &\geq& C' \frac{r \n^4 \log(\n/\delta)}{\qmin \sigma_1(\M_2)^2\varepsilon^2} \Big(\frac{1}{\l^2} +\frac{\sigma_1(\M_2)}{\l \n}+\frac{r^4\sigma_1(\M_2)^4}{\sigma_r(\M_2)^5}\Big)
	\;,\text{ and }\\
		\n &\geq& C' {\r^{3.5}\mu^6} \,\Big(\frac{\sigma_1(\M_2)}{\sigma_r(\M_2)}\Big)^{4.5}\;,
	\end{eqnarray*} 
	then there exists a permutation $\pi$ over $[\r]$ such that Algorithm \ref{algo:spect} 
	achieves the following bounds with a choice of $T=C'\log(\n\,|\cS|)$ 
	for all $i\in[r]$, with probability at least $1-\delta$:
	\begin{eqnarray*}
		| \hq_{\pi_i} - \q_i| \;\leq\; \eps \;,\text{ and } \;\;\;\;\;
		\| \hP_{\pi_i} - \P_i \| \;\leq\; \eps\,\sqrt{\frac{r\,\qmax\,\sigma_1(\M_2)}{\qmin}}\;,
	\end{eqnarray*}
	where $\mu=\mu(\M_2)$ defined in \eqref{eq:defmu} with 
	run-time  ${\rm poly}(\n,\r,1/\qmin,1/\eps,\log(1/\delta),\sigma_1(\M_2)/\sigma_\r(\M_2))$.
\end{thm}

\begin{algorithm}[bht]
\caption{\matrixcompletion: Alternating Minimization for Matrix Completion \cite{JO14}}
\label{algo:altmin}
\begin{algorithmic}[1]
\STATE {\bf Input}: $\hat{M}_2\big(1,\frac{|\cS|}{2}\big)$, $r$, $T$
\STATE Initialize $\n\times\r$ dimensional matrix $U_0\leftarrow $ top-$r$ eigenvectors of $\cP_{\Omega_2}(\hat{M}_2\big(1,\frac{|\cS|}{2}\big))$
\FORALL{$\tau=1 $ to $T-1$}
\STATE $\hU_{\tau+1}=\arg\min_{U}\|\cP_{\Omega_2}(\hat{M}_2\big(1,\frac{|\cS|}{2}\big))-\cP_{\Omega_2}(UU_\tau^T)\|_F^2$
\STATE $[U_{\tau+1} R_{\tau+1}]={\rm QR}(\hU_{\tau+1})$  \hfill (standard QR decomposition)
\ENDFOR
\STATE {\bf Output}: $\tM_2=(\hU_{T})(U_{T-1})^T$
\end{algorithmic}
\end{algorithm}

\begin{algorithm}[bth]
\caption{\tensorLS: Least Squares method for Tensor Estimation \cite{JO14}}
\label{algo:ls}
\begin{algorithmic}[1]
\STATE {\bf Input}: $\hat{M}_3\big(\frac{|\cS|}{2}+1, |\cS|\big)$, $\hU_{M_2}$, $\hsm$
\STATE Define operator $\hnu: \reals^{\r\times \r\times \r}\rightarrow \reals^{\n\times\n\times\n}$ as follows
\begin{equation}
  \label{eq:hnu}
  \hnu_{ijk}(Z)=\begin{cases}\sum_{abc}Z_{abc}(\hU_{M_2}\hsm^{1/2})_{ia}(\hU_{M_2}\hsm^{1/2})_{jb}(\hU_{M_2}\hsm^{1/2})_{kc},& \text{ if } i\neq j\neq k\neq i\;, \\
0, & \mbox{otherwise}.\end{cases}
\end{equation}
\STATE Define $\hA: \reals^{r\times r\times r}\rightarrow \reals^{r\times r\times r}$ s.t. $\hA(Z)=\hnu(Z)[\hU_{M_2}\hsm^{-1/2}, \hU_{M_2}\hsm^{-1/2}, \hU_{M_2}\hsm^{-1/2}]$
\STATE {\bf Output}: $\arg\min_{Z} \|\hA(Z)-\cP_{\Omega_3}\big(\hat{M}_3\big(\frac{|\cS|}{2}+1, |\cS|\big)\big)[\hU_{M_2}\hsm^{-1/2}, \hU_{M_2}\hsm^{-1/2}, \hU_{M_2}\hsm^{-1/2}]\|_F^2$
\end{algorithmic}
\end{algorithm}

\subsection{Phase 2: {\sc RankCentrality}.}\label{algo:rank}

Recall that $E = \{(i_k, j_k): i_k \neq j_k \in [n], 1\leq k \leq N\}$ represents collection of $N=|E|$ pairs and $G = ([n], E)$ is the corresponding graph.  Let $\tP_a$ 
denote the estimation of $P_a=[P_{ka}] \in [-1,1]^N$ for the mixture component $a, 1\leq a\leq r$; where $P_{ka}$ is 
defined as per \eqref{eq:defP}. For each $a$, using $G$ and $\tP_a$, we shall use the {\sc RankCentrality} \cite{NOS12} 
to obtain estimation of $\bw^{(a)}$.  Next we describe the algorithm and guarantees associated with it.

Without loss of generality, we can assume that $\bw^{(a)}$ is such that $\sum_i w^{(a)}_i = 1$ for all $a, 1\leq a \leq r$. Given 
this normalization, {\sc RankCentrality} estimates $\bw^{(a)}$ as stationary distribution of an appropriate Markov chain
on $G$. The transition probabilities are $0$ for all $(i,j) \notin E$. For $(i,j) \in E$, they are function of $\tP_a$. Specifically,  
transition matrix $\tp^{(a)} = [\tp_{i,j}^{(a)}] \in [0,1]^{n\times n}$ with $\tp_{i,j}^{(a)} = 0$ if $(i,j) \notin E$, and for $(i_k, j_k) \in E$ for $1\leq k\leq N$, 
\begin{align}
\tp_{i_k,j_k}^{(a)} &=  
	 \frac{1}{\dmax} \frac{(1+\tP_{ka})}{2} \quad \text{ and } \quad \tp_{j_k,i_k}^{(a)}~=~ \frac{1}{\dmax} \frac{(1-\tP_{ka})}{2}, \label{eq:defrw}
\end{align}
Finally, $\tp_{i,i}^{(a)} = 1-\sum_{j\neq i} \tp_{i,j}^{(a)}$ for all $i \in [n]$. Let $\tpi^{(a)} = [\tpi_i^{(a)}]$ be a stationary
distribution of the Markov chain defined by $\tp^{(a)}$. That is, 
%
\begin{align}
	\tpi_i^{(a)} &= \sum_j \,\tp_{ji}^{(a)}\,\tpi_j^{(a)}\quad \mbox{for all}~i\in [n].
\end{align}
Computationally, we suggest obtaining estimation of $\tpi$ by using power-iteration for $T$ iterations. As argued before, cf. \cite{NOS12}, 
$T=\Theta\big(\,\b^2 \dmax(\log n + \log(1/\varepsilon))/(\xi\dmin)\,  \big)$, is sufficient to obtain reasonably good estimation of $\tpi$. 

The underlying assumption here is that there is a unique stationary distribution, which is established by our result under the conditions of
Theorem \ref{thm:main}. Now $\tp$ is an approximation of the ideal transition probabilities, where $p^{(a)} = [p_{i,j}^{(a)}]$ 
where $p^{(a)}_{i,j} = 0$ if $(i,j) \notin E$ and $p^{(a)}_{i,j} \propto w_j^{(a)}/(w_i^{(a)} + w_j^{(a)})$ for all $(i,j) \in E$. Such an ideal
Markov chain is {\em reversible} and as long as $G$ is connected (which is, in our case, by choice), the stationary distribution
of this ideal chain is $\pi^{(a)} = \bw^{(a)}$ (recall, we have assumed $\bw^{(a)}$ to be normalized so that all its components up to $1$). 

Now $\tp^{(a)}$ is an approximation of such an ideal transition matrix $p^{(a)}$. In what follows, we state result about how this approximation error
translates into the error between $\tpi^{(a)}$ and $\bw^{(a)}$.  Recall that $\b \equiv \max_{i,j\in [n]} w_i/\w_j$, $\dmax$ and $\dmin$
are maximum and minimum vertex degrees of $G$ and $\xi$ as defined in \eqref{eq:defxi}.  
\begin{thm}
\label{thm:rankcentrality}	
Let $G = ([n], E)$ be non-bipartite and connected. Let $\|\tp^{(a)}-p^{(a)}\|_2 \leq \varepsilon$ for some positive 
$\varepsilon\leq (1/4)\xi\b^{-5/2}(\dmin/\dmax)$. Then, for some positive universal constant $C$,
\begin{align}
	\frac{\|\tpi^{(a)} -\bw^{(a)}\|}{\|\bw^{(a)}\|} &\leq \frac{C\,\b^{5/2} }{\xi} \frac{\dmax}{\dmin}\,\varepsilon.
\end{align}
And, starting from any initial condition, the power iteration manages to produce an estimate of $\tpi^{(a)}$ within twice the above stated error bound
in $T=\Theta\big(\,\b^2 \dmax(\log n + \log(1/\varepsilon))/(\xi\dmin)\,  \big)$ iterations. 
\end{thm}
Proof of the above result can be found in Appendix. For spectral expander (e.g. connected Erdos-Renyi graph with high probability), $\xi = \Theta(1)$ 
and therefore the bound is effectively $O(\varepsilon)$ for bounded dynamic range, i.e. $b= O(1)$.

\section{Discussion}

Learning distribution over permutations of $n$ objects from partial observation is  fundamental to many
domains. In this work, we have advanced understanding of this question by characterizing sufficient 
conditions and associated algorithm under which it is feasible to learn mixed MNL model in computationally 
and statistically efficient (polynomial in problem size) manner from partial/pair-wise comparisons. 
The conditions are natural -- the mixture components should be ``identifiable'' given partial preference/comparison 
data -- stated in terms of full rank and incoherence conditions of the second moment matrix. The algorithm 
allows learning of mixture components as long as number of mixture components scale $o(n^{2/7})$ for 
distribution over permutations of $n$ objects.  

To the best of our knowledge, this work provides first such sufficient condition for learning mixed MNL model -- 
a problem that has remained open in econometrics and statistics for a while, and 
more recently Machine learning. Our work nicely complements the impossibility results of \cite{AOSV14}. 

Analytically, our work advances the recently popularized spectral/tensor approach for learning mixture model from
lower order moments. Concretely, we provide means to learn the component even when only partial information
about the sample is available unlike the prior works. To learn the model parameters, once we identify the moments
associated with each mixture, we advance the result of \cite{NOS12} in its applicability. 
Spectral methods have also been applied to ranking in the context of assortment optimization in \cite{BGG13}.


{\small
\bibliographystyle{plain}
\bibliography{ranking}
}

\newpage
{\Large{\bf Supplementary material for ``Learning Mixed Multinomial Logit Model from Ordinal Data''}}
\appendix

\section{Proof of Theorem \ref{thm:main}}
\label{sec:proofmain}

In order to apply Theorem \ref{thm:rankcentrality}, 
let $\Delta=p - \tp$, then 
\begin{align*}
	\| \Delta\|_2 &\leq  \|\Delta-\diag(\Delta)\|_2 \;+\; \|\diag(\Delta)\|_2 
		~\leq \|\Delta-\diag(\Delta)\|_F \;+\; \max_{i\in[n]} \Delta_{ii} \\
		&\leq \sqrt{\sum_{i\neq j} (p_{ij}-\tp_{ij})^2} \;+\; \max_{i\in[n]} \Big| \sum_{j\neq i} (p_{ij}-\tp_{ij})\Big| \\
		& \leq \frac{1}{\sqrt{2}\,\dmax} \|\P_a-\tP_a\| \;+\; \frac{1}{2\sqrt{\dmax}} \|\P_a-\tP_a\|\;.
\end{align*}
From Theorem \ref{thm:finite}, we know that 
$\|\Delta\|_2\leq \frac{2}{\sqrt{\dmax}} \|\P_a-\tP_a\| \leq \varepsilon \sqrt{(\r \qmax \sigma_1(\M_2))/\qmin}$ 
and substituting this into the bound in Theorem \ref{thm:rankcentrality} we get the desired bound. 

%
%
\section{Proof of the performance guarantee for the spectral method in Theorem~\ref{thm:finite}}

To simplify notations, we will assume that the indices of the output of the algorithm and the ground truths are matched such that the theorem holds with identity permutation. 
The spectral algorithm outputs $\hP = \tU_{\M_2} \tSigma_{\M_2}^{1/2} \hV^{\tH} \hLambda^{\tH}$. 
From theorem \ref{thm:consistency}, we know that $\P = \U_{\M_2}\Sigma_{\M_2}^{1/2}\V^\G \Lambda^\G$, or equivalently 
$\P = \U_{\M_2}\Sigma_{\M_2}^{1/2}\V^\G Q^{-1/2}$. 
To show that $\P$ and $\hP$ are close, we would hope that each of the terms above to be close. 

%
%

To that end, define 
\begin{eqnarray}
	\tV &\equiv& \tSigma_{M_2}^{-1/2} \tU_{M_2}^T \P \Q^{1/2}\;,\label{eq:deftV}\\ 
	\tG &\equiv& \sum_{i=1}^\r \frac{1}{\sqrt{\q_i}} (\tv_i\otimes \tv_i\otimes \tv_i)\;,\label{eq:deftG}
\end{eqnarray}
where $\tV=[\tv_1\,\tv_2\,\ldots \tv_\r]$ (note that $\tG$ is proxy of $V^\G$). 
Now
\begin{eqnarray}
  \big\|\, \hP-\P\, \big\|_2 &\leq& 
  	\big\|\, \tU_{M_2}\tU_{M_2}^T\P - \P\, \big\|_2 + \big\|\, \hP- \tU_{M_2}\tU_{M_2}^T\P\, \big\|_2   \nonumber\\ 
	&=&	\big\|\, \tU_{M_2}\tU_{M_2}^T\P - \P\, \big\|_2 + \big\|\, \tU_{\M_2} \tSigma_{\M_2}^{1/2} \hV^{\tH} \hLambda^{\tH} - \tU_{M_2} \tSigma_{M_2}^{1/2} \tV \Q^{-1/2}\, \big\|_2 \qquad \text{using~}\eqref{eq:deftV}  \nonumber\\ 
	&\leq& \big\|\, \tU_{M_2}\tU_{M_2}^T\P - \P\, \big\|_2 + \big\| \,\tU_{\M_2} \tSigma_{\M_2}^{1/2}\,( \tV-\hV^{\tH})\,Q^{-1/2}\big\|_2 \nonumber \\
	& ~ & \qquad + \big\|\tU_{\M_2} \tSigma_{\M_2}^{1/2}\hV^{\tH}\,(Q^{-1/2}-\hLambda^{\tH})\, \big\|_2 \label{eq:errorbound}\;.
\end{eqnarray} 
To bound the three terms on the RHS of \eqref{eq:errorbound}, we shall use the following `errors' (which we shall 
bound sharply later in the proof): define (recall $\tH$ was produced by Algorithm  Tensor Least Squares)
\begin{align}
	\eps_{\M} &= \frac{\| \M_2 - \tM_2\|_2}{\sigma_r(\M_2)} \nonumber \\
	 \eps_\G & = \|\tH-\G\|_2. \label{eq:errordef}
\end{align} 
\noindent{\bf Bounding the first term in RHS of  \eqref{eq:errorbound}.}
To begin with, note that we can represent $P = U_{M_2} \Sigma V^T$ by definition. Given the definition of \eqref{eq:errordef},
and an application of Davis-Kahan theorem \cite{DK70} implies that 
\begin{align}\label{rem:M2error}
\|( \tU_{M_2}\tU_{M_2}^T -\id )U_{M_2} \|_2 &\leq \eps_{M}. 
\end{align}
Using this, we have 
\begin{align}
	\|\tU_{M_2}\tU_{M_2}^T\P - \P\|_2 & = \|(\tU_{M_2}\tU_{M_2}^T -\id) \P \|_2 \nonumber \\
	       & \leq \|(\tU_{M_2}\tU_{M_2}^T -\id) \|_2 U_{M_2} \| \Sigma V^T \|_2 \nonumber \\
	       & \leq \eps_{\M}\|P\|_2 \;,\label{eq:errorbound1}
\end{align}
\noindent{\bf Bounding the second term in RHS of  \eqref{eq:errorbound}.} 
Consider 
\begin{align}
\|\tSigma_{\M_2}\|_2 & = \|\tM_2\|_2 \leq \|\tM_2-\M_2\|_2 + \|\M_2\|_2 \nonumber \\
                                     & \leq \eps_{M}\sigma_r(\M_2) + \|M_2\|_2. \label{eq:z1}
\end{align} 
We state the following Lemma providing bound on $\|\tV-\hV^{\tH}\|_2$ in terms of $\eps_M$ and $\eps_G$:
\begin{lemma}
	\label{lem:powerbound}
	There exists universal constant $C_1 > 0$ such that 
	\begin{eqnarray*}
		\| \tV-\hV^{\tH}\|_2 &\leq& C_1\sqrt{r\,\qmax}\,\Big(\,\eps_\G + \frac{1}{\sqrt{\qmin}} \,\eps_\M\,\Big) \;, \text{ and }\\
		\| Q^{-1/2}-\hLambda^\G\|_2 &\leq& C_1\Big(\,\eps_\G+\frac{1}{\sqrt{\qmin}}\,\eps_\M\,\Big) \;.
	\end{eqnarray*}
\end{lemma}
Given \eqref{eq:z1}, Lemma \ref{lem:powerbound}, the fact that $\tU_{\M_2}$ is a unitary matrix and $Q$ is a diagonal matrix, we obtain that the 
second term in RHS of \eqref{eq:errorbound} is bounded by 
\begin{eqnarray}
	\big\| \,\tU_{\M_2} \tSigma_{\M_2}^{1/2}\,( \tV-\hV^{\tH})\,Q^{-1/2}\big\|_2  & \leq & \big\| \,\tU_{\M_2} \big\|_2\big\|  \tSigma_{\M_2}^{1/2}\, \big\|_2\big\|  ( \tV-\hV^{\tH})\, \big\|_2 \big\| Q^{-1/2}\big\|_2\nonumber\\
	&\leq& 
		\frac{\sqrt{\eps_{M}\sigma_r(\M_2) + \|M_2\|_2}}{\sqrt{\qmin}} \, \|\tV-\hV^{\tH}\|_2 \nonumber\\
		&\leq& C_2\,  \sqrt{\frac{\|\M_2\|_2\,\r\,\qmax}{\qmin}}\Big( \eps_\G + \frac{1}{\sqrt{\qmin}} \,\eps_\M\Big)   \;, \label{eq:errorbound2}
\end{eqnarray}
for an appropriate universal constant $C_2>0$, with $\eps_M\leq1/2$ and using fact $\sigma_r(M_2) \leq \|M_2\|_2$. 


\noindent{\bf Bounding the third term in RHS of \eqref{eq:errorbound}.} Observe that
\begin{align}\label{eq:z2}
\|\hV^{\tH}\|_2 & \leq \|\tV\|_2+\|\hV^{\tH}-\tV\|_2 \nonumber \\
                     & \leq 1+2\eps_\M+ C_1 \sqrt{r\qmax}(\eps_\G+\eps_\M/\sqrt{\qmin})
\end{align}
using Remark \ref{rem:boundbV} and Lemma \ref{lem:powerbound}. Using \eqref{eq:z2} and 
Lemma \ref{lem:powerbound}, the last term in \eqref{eq:errorbound} can be bounded above as 
\begin{align}
	& \big\|\tU_{\M_2} \tSigma_{\M_2}^{1/2}\hV^{\tH}\,(Q^{-1/2}-\hLambda^{\tH})\, \big\|_2 ~\leq~\big\| \tU_{\M_2} \tSigma_{\M_2}^{1/2} \big\|_2 \big\| \hV^{\tH}\, \big\|_2 \big\| (Q^{-1/2}-\hLambda^{\tH})\, \big\|_2  \nonumber \\
	& \quad \leq \big(\sqrt{\eps_{M}\sigma_r(\M_2) + \|M_2\|_2}\big) \, \|\hV^{\tH}\|_2\, \| Q^{-1/2}-\hLambda^G \|_2 \nonumber\\
		& \quad \leq C_3 \sqrt{\|\M_2\|_2}\,  \Big(\,1 + \sqrt{r\qmax}\eps_\G+\frac{\sqrt{r\qmax}}{\sqrt{\qmin}}\,\eps_\M\,\Big)\, \Big(\,\eps_\G+\frac{1}{\sqrt{\qmin}}\,\eps_\M\,\Big) \\
		& \quad \leq C_3 \sqrt{\|\M_2\|_2} \, \Big(\,\eps_\G+\frac{1}{\sqrt{\qmin}}\,\eps_\M\,\Big) \;,
	\label{eq:errorbound3}
\end{align}
for $\eps_\M\leq \sqrt{\qmin/(\r\,\qmax)}$, $\eps_\G\leq1/\sqrt{r\,\qmax}$ and for some universal constant $C_3>0$.
%
%
%
%

\noindent{\bf Towards Theorem \ref{thm:finite}.}
Substituting \eqref{eq:errorbound1}, \eqref{eq:errorbound2}, \eqref{eq:errorbound3} in \eqref{eq:errorbound}, for universal constant $C_4> 0$, we get
\begin{align}\label{eq:z4}
	\|\hP-\P\|_2 &\leq \eps_\M \|P\|_2 + C_4 \,\sqrt{\frac{\|\M_2\|_2\,\r\,\qmax}{\qmin}}\Big(\,\eps_\G+\frac{1}{\sqrt{\qmin}}\,\eps_\M\,\Big) \nonumber \\
		&\leq C_4 \,\sqrt{\frac{\|\M_2\|_2\,\r\,\qmax}{\qmin}}\Big(\,\eps_\G+\frac{1}{\sqrt{\qmin}}\,\eps_\M\,\Big) \;, 
\end{align}
where we used the fact that $\|\P\|_2\leq\|\sqrt{\|\M_2\|_2/\qmin}\|$ since $\P = \U_{\M_2}\Sigma_{\M_2}^{1/2}\V^\G \Lambda^\G$. 
Given \eqref{eq:z4}, to complete  the proof of Theorem ~\ref{thm:finite}, we need to establish bounds on $\eps_M$ and $\eps_\G$. 


\noindent{\bf Bounding $\eps_M$.} To bound $\eps_M$, we need to bound error between $M_2$ and $\tM_2$, the
output of alternating minimization procedure applied to $\hM_2$. The following theorem \cite{JO14} provides such
a bound.
\begin{thm}[Theorem 4.1,\,\cite{JO14}]
	\label{thm:M2complete}
	For an $\n \times \n$ symmetric rank-$\r$ matrix $\M$ with incoherence $\mu$, 
	we observe off-diagonal entries corrupted by noise: \vspace*{-2pt}
	\begin{eqnarray*}
		\hM_{ij} &=& \left\{ \begin{array}{rl} \M_{ij} + E_{ij} & \text{ if }i\neq j\;, \\ 
			0 & \text{ otherwise.}\end{array} \right.\vspace*{-15pt}
	\end{eqnarray*}
	Let $\hM^{(\tau)}$ denote the output after $\tau$ iterations of \matrixcompletion. 
	If $\mu \leq (\sigma_\r(\M)/\sigma_1(\M)) \sqrt{\n/(32\,\r^{1.5})} $, 
	the noise is bounded by 
	$\|\cP_{\Omega_2}(E)\|_2 \leq \sigma_\r(\M)  /32\sqrt{\r}$, 
	and each column of the noise is bounded by 
	$\|\cP_{\Omega_2}(E)_{i}\| \leq \sigma_1(\M) \mu \sqrt{3\r/(8\,\n)}$, $\forall i\in[\n]$,  
	then after $\tau \geq (1/2)\log\big(2\|\M\|_F/\varepsilon \big)$ iterations of \matrixcompletion, 
	the estimate $\hM^{(\tau)}$ satisfies:\vspace*{-5pt} 
	\begin{eqnarray*}
		\| \M - \hM^{(\tau)} \|_2 &\leq& \varepsilon +  \frac{9\, \|\M\|_F \,\sqrt{\r}\,}{\sigma_\r(\M)}\|\cP_{\Omega_2}(E)\|_2 \;, \vspace*{-5pt}
	\end{eqnarray*}
	for any $\varepsilon\in(0,1)$. 
	Further, $\hM^{(\tau)}$ is $\mu_1$-incoherent with $\mu_1\leq 6\mu\sigma_1(\M_2)/\sigma_\r(\M_2)$. 
\end{thm}
To apply the above result in our setting to bound $\eps_M$, we need to bound $\| \cP_{\Omega_2}(E) \|_2$
and $\| \cP_{\Omega_2}(E)_i\|$ for all $i \in [N]$. To that end, we state the following Lemma. 
\begin{lemma}
	\label{lem:M2sample}
	Let $S_2 \equiv (2/|\cS|) \sum_{t\in\{1,\ldots,|\cS|/2\}} x_t x_t^T $ 
	be the sample covariance matrix, and let 
	$E=\frac{\n(\n-1)}{\l(\l-1)}S_2-M_2$ denote the sampling error in the off-diagonal entries. Then, 
	there exists a universal  constant $C_5 >0$ such that with probability at least $1-\delta$, 
	\begin{eqnarray*}
		\| \cP_{\Omega_2}(E) \|_2 &\leq& C_5 \sqrt{\frac{ \n^2 \log(\n/\delta) }{\l|\cS|} \Big(\sigma_1(\M_2)+\frac\n\l\Big)}\;.
	\end{eqnarray*}
	Moreover, the Euclidean norm of the columns are uniformly bounded by 
	\begin{eqnarray*} 
		\| \cP_{\Omega_2}(E)_i\| &\leq& C_5 \sqrt{\frac{\n^3 \log(\n/\delta) }{\l^2 |\cS|}}\;, 
	\end{eqnarray*}
	for all $i\in[\n]$. 
\end{lemma}
Theorem \ref{thm:M2complete} and Lemma \ref{lem:M2sample} imply that with probability at least $1-\delta$, for large enough iterations 
of the \matrixcompletion, 
\begin{align}\label{eq:z5}
	\eps_M &= \frac{1}{\sigma_\r(\M_2)} \|\M_2-\tM_2\|_2 \;\leq\; C_6 \frac{\|\M_2\|_F\, \n}{\sigma_r(\M_2)^2}\sqrt{\frac{\r \, \log(\n/\delta)}{\l\,|\cS|}\Big(\sigma_1(\M_2)+\frac\n\l\Big)} \;,
\end{align}
for universal constant $C_6, C_7 > 0$ when $|\cS| \geq C_7 \n^4 \r \log(\n/\delta) /( \l^2 \sigma_\r(\M_2)^2)$ -- the assumption of Theorem
statement. 

\noindent{\bf Bounding $\eps_\G$.} To bound $\eps_\G$, we need bound on error induced
by the output of the Tensor Least Square procedure. The following result \cite{JO14} provides such
a bound.
\begin{thm}[Theorem 4.3,\,\cite{JO14}]
	\label{thm:M3LS}
	If $\n\geq\frac{144 \r^3\sigma_1(\M_2)^2}{\sigma_r(\M_2)^2}$, then with probability at least $1-\delta$, 
	\begin{eqnarray*}
		\|\G-\hG\|_F &\leq& \frac{24 \mu_1^3 \mu \r^{3.5} \sigma_1(\M_2)^{3/2}}{\n\qmin^{1/2}\sigma_\r(\M_2)^{3/2}} \,\eps_\M + 2 \Big\| \, \cP_{\Omega_3}\big(S_3-\M_3\big)\big[\tU_{\M_2}\tSigma_{\M_2}^{-1/2},\tU_{\M_2}\tSigma_{\M_2}^{-1/2},\tU_{\M_2}\tSigma_{\M_2}^{-1/2}\big]\, \Big\|_F \;, 
	\end{eqnarray*}
	where $\eps_\M=(1/\sigma_r(\M_2)) \|\tM_2-\M_2\|$, $\mu=\mu(M_2)$, $\mu_1=\mu(\tU_{M_2})$, and
\begin{align}
S_3 & =\frac{\n(\n-1)(\n-2)}{\l(\l-1)(\l-2)}\frac{2}{|\cS|} \Big(\sum_{t=1+|\cS|/2}^{|\cS|}x_t\otimes x_t\otimes x_t\Big).
\end{align}
\end{thm}
To utilize above result, we state the following Lemma. 
\begin{lemma}
	\label{lem:M3sample}
	There exists a positive numerical constant $C_9$ such that with probability at least $1-\delta$, 
	\begin{eqnarray*}
	\Big\| \, \cP_{\Omega_3}\big(S_3-\M_3\Big)\big[\tU_{\M_2}\tSigma_{\M_2}^{-1/2},\tU_{\M_2}\tSigma_{\M_2}^{-1/2},\tU_{\M_2}\tSigma_{\M_2}^{-1/2}\big]\, \big\|_F
	&\leq& C_9\frac{\r^3\mu_1^3\n^{3/2}}{\sigma_r(\M_2)^{3/2}}\sqrt{\frac{\log(1/\delta)}{|\cS|}}\;.
	\end{eqnarray*}
\end{lemma}
Theorem \ref{thm:M3LS} and Lemma \ref{lem:M3sample} imply that with probability at least $1-\delta$
\begin{align}\label{eq:z6}
	\eps_\G &= \|\G-\tH\|_2 \nonumber \\
	& \leq  C_{10}\, \mu_1^3\, \r^3\, \left(\,\frac{\mu \r^{1/2}}{\n\qmin^{1/2}}\frac{\sigma_1(\M_2)^{3/2}}{\sigma_r(\M_2)^{3/2}} \eps_\M + \frac{\n^{3/2}}{\sigma_r(\M_2)^{3/2}}\sqrt{\frac{\log(1/\delta)}{|\cS|}}\,\right)
\end{align}
for some positive numerical constants $C_{10}, C_{11}$ when $\n\geq C_{11} \frac{\r^3\sigma_1(\M_2)^2}{\sigma_r(\M_2)^2}$. 

Equations \eqref{eq:z4}-\eqref{eq:z6} imply that 
\begin{align}
	& \|\hP-\P\|_2 \leq C_4 \,\sqrt{\frac{\|\M_2\|_2\,\r\,\qmax}{\qmin}}\Big(\,\eps_\G+\frac{1}{\sqrt{\qmin}}\,\eps_\M\,\Big) \;, \nonumber \\
	& \leq C_4 \,\sqrt{\frac{\|\M_2\|_2\,\r\,\qmax}{\qmin}}\Big(\, C_{10}\, \mu_1^3\, \r^3\, \left(\,\frac{\mu \r^{1/2}}{\n\qmin^{1/2}}\frac{\sigma_1(\M_2)^{3/2}}{\sigma_r(\M_2)^{3/2}} \eps_\M + \frac{\n^{3/2}}{\sigma_r(\M_2)^{3/2}}\sqrt{\frac{\log(1/\delta)}{|\cS|}}\,\right) +\frac{1}{\sqrt{\qmin}}\,\eps_\M\,\Big) \;, \nonumber.
\end{align}
Using $\mu_1\leq 6\mu\sigma_1(\M_2)/\sigma_\r(\M_2)$ from Theorem \ref{thm:M2complete}, we obtain (for appropriate constant $C_{12} > 0$), 
\begin{align}\label{eq:z7}	
\|\hP-\P\|_2 & \leq C_{12} \,\sqrt{\frac{\|\M_2\|_2\,\r\,\qmax}{\qmin^2}}  \left\{ \Big(\frac{\mu^4 \r^{3.5}}{\n}\Big(\frac{\sigma_1(\M_2)}{\sigma_r(\M_2)}\Big)^{4.5} + 1 \Big) \eps_\M + \frac{\mu^3 r^3 \sigma_1(\M_2)^3\n^{3/2} \qmin^{1/2}}{\sigma_r(\M_2)^{4.5}}\sqrt{\frac{\log(1/\delta)}{|\cS|}}\, \right\}. \nonumber \\
&\leq C_{12} \sqrt{\frac{\|\M_2\|_2\,\r\,\qmax}{\qmin}} \left\{ \frac{2\eps_\M}{\qmin^{1/2}} \,+\, \mu^3 \r^3 \frac{\sigma_1(\M_2)^3\n^{1.5}}{\sigma_r(\M_2)^{4.5}}\sqrt{\frac{\log(1/\delta)}{|\cS|}} \right\}\;, 
\end{align}
where $\n\geq \mu^4 \r^{3.5} (\sigma_1(\M_2)/\sigma_\r(\M_2))^{4.5}$ as per assumption of the Theorem statement. 
From \eqref{eq:z5}, it follows that when 
\begin{align}\label{eq:z8}
	|\cS| &\geq C_{13} \frac{\r^2 \sigma_1(M_2)^2 N^2(\sigma_1(\M_2)+\n/\l) \log(N/\delta)}{\varepsilon^2 \qmin \ell \sigma_r(\M_2)^4} 
	~~\geq C_{13} \frac{\r \|M_2\|_F^2 N^2 \log(N/\delta)}{\varepsilon^2 \qmin \ell \sigma_r(\M_2)^4}\Big(\sigma_1(\M_2)+\frac\n\l\Big),
\end{align}
(we used $\|M_2\|_F \leq \sqrt{r} \sigma_1(\M_2)$ for rank $r$ matrix $\M_2$) 
for an appropriate choice of universal constant $C_{13} > 0$,
\begin{align}
\frac{2\eps_\M}{\qmin^{1/2}} & \leq \frac{\varepsilon}{2}.
\end{align}
Also, recall that for \eqref{eq:z5} to hold, we require
\begin{eqnarray}
 |\cS| \geq C_7 \frac{\n^4 \r \log(\n/\delta)}{ \l^2 \sigma_\r(\M_2)^2}\;.
\end{eqnarray}
Also, when 
\begin{align}\label{eq:z8}
|\cS| &\geq C_{14} \frac{\mu^6 \r^6 N^3 \sigma_1(\M_2)^6\log(1/\delta) }{\varepsilon^2 \sigma_r(\M_2)^9},
\end{align}
for an appropriate choice of universal constant $C_{14} > 0$, the second term inside bracket in \eqref{eq:z7} is less than $\frac{\varepsilon}{2}$. From above, it
follows that when 
\begin{align}
	|\cS| &\geq C_{15}  \frac{\r^2 N^2 \sigma_1(\M_2)^2 \log(N/\delta)}{\qmin \varepsilon^2 \sigma_\r(\M_2)^4} \Big(\frac{N}{ \l^2} + \frac{\sigma_1(\M_2)}{\l} + \frac{\qmin \mu^6 \r^4 \sigma_1(\M_2)^4\n}{\sigma_\r(\M_2)^5}\Big) 
	+ \frac{\n^4r\log(\n/\delta)}{\l^2\sigma_1(\M_2)^2}\, 
\end{align}
\begin{eqnarray*}
	 \|\hP-\P\|_2 &\leq& \eps \, \sqrt{\frac{\r\,\qmax\sigma_1(\M_2)}{\qmin}} \;, \text{ and }\\
	 \|\Q^{-1/2}-\hLambda^G\|_2 &\leq& \eps \;,
\end{eqnarray*}
for any $\eps\in (0,C')$ for some positive constants $C = C_{15}$ and $C'$. 
Assuming $\n\geq C'' r^{3.5} \mu^6 (\sigma_1(\M_2)/\sigma_r(\M_2))^{4.5}$, 
the above holds for  
\begin{eqnarray*}
	|\cS| &\geq& C'''\frac{r \n^4 \log(\n/\delta)}{\qmin\sigma_1(\M_2)^2\varepsilon^2} \Big(\frac{1}{\l^2} +\frac{\sigma_1(\M_2)}{\l \n}+\frac{r^4\sigma_1(\M_2)^4}{\sigma_r(\M_2)^5}\Big)\;.
\end{eqnarray*}

\section{Proof of the technical lemmas for the spectral method}


\subsection{Proof of Lemma \ref{lem:powerbound}}

In order to apply the perturbation analysis of Theorem \ref{thm:aghkt} from \cite{AGH12}, 
it is crucial that we compare to a tensor with an orthogonal decomposition. 
Since both $\hG$ and $\tG$ do not have orthogonal decompositions, we define a new tensor 
$\bG$ that is close to $\tG$ and has an orthogonal decomposition. 
Given the singular value decomposition of $\tV=XSY^T$, 
define $\bV\equiv XY^T$. 
This $\bV\in\reals^{\r\times\r}$ is orthogonal such that $\bV\bV^T=\bV^T\bV=\id$, and is close to $\tV$ such that 
\begin{eqnarray}
	\|\tV-\bV\|_2 &=& \| X(S-\id)Y^T\|_2 \nonumber\\
	 	&\leq& \max_{i\in[r]} |S_{ii}-1|\nonumber\\
		&\leq& 2\eps_M\;, \label{eq:Vbound}
\end{eqnarray}
where the last inequality follows from the next remark. 
\begin{remark} [Remark 10 in \cite{JO14}]
	Suppose $\|\M_2-\tM_2\|_2\leq \eps_\M \sigma_r(\M_2)$, then 
	$$ \|\id - \tV\tV^T\|_2 \;\leq\; 2\eps_\M\;.$$
	\label{rem:boundbV}
\end{remark}
It follows that $\|S^2-\id\|_2 = \|\tV\tV^T-\id\|_2\leq2\eps_\M$. 
Therefore, $S_{ii}^2\in[1-2\eps_\M,1+2\eps_\M]$ and so is 
$S_{ii}\in[1-2\eps_\M,1+2\eps_\M]$ for $\eps_\M\leq1/2$.

Since $\|\tV-\hV^{\tH}  \|_2 \leq \| \tV-\bV \|_2 + \| \bV-\hV^{\tH}  \|_2  
		\leq 2\eps_\M +  \| \bV-\hV^{\tH}  \|_2$, 
we are left to show that $ \| \bV-\hV^{\tH} \|_2\leq8 \sqrt{\r\,\qmax} (\eps_\G+(13/\sqrt{\qmin})\eps_\M)$ to finish the proof. 
Recall that $\tG= \sum_{i=1}^\r \frac{1}{\sqrt{\q_i}}(\tv_i\otimes \tv_i\otimes \tv_i)$ 
and that $\hV^{\tH} $ is the output of the robust power method applied to $\hG$, and let  
\begin{eqnarray*}
	\bG &\equiv& \sum_{i=1}^\r \frac{1}{\sqrt{\q_i}}(\bv_i\otimes \bv_i\otimes \bv_i)\;.	
\end{eqnarray*}

Applying \eqref{eq:Vbound}, we get that 
\begin{eqnarray*}
	\|\tG-\bG\|_2 &=& \max_{\|u\|=1} \Big\| \sum_{i=1}^\r \frac{1}{\sqrt{\q_i}} (\tv_i\otimes \tv_i\otimes \tv_i-\bv_i\otimes \bv_i\otimes \bv_i)\Big\|_2\\
		&\leq& \max_{\|u\|=1}\, \sum_{i=1}^\r \frac{1}{\sqrt{\q_i}} \Big\{(u^T\tv_i)^3-(u^T\bv_i)^3 \Big\} \\
		&\leq& \max_{\|u\|=1}\, \sum_{i=1}^\r \frac{1}{\sqrt{\q_i}} \Big(u^T(\tV-\bV)e_i\Big) \Big((u^T\tv_i)^2+(u^T\tv_i)(u^T\bv_i)+(u^T\bv_i)^2 \Big) \\
		&\leq& \frac{1}{\sqrt{\qmin}}\|\tV-\bV\|_2 (3+6\eps_\M+\eps_\M^2)\\
		&\leq& \frac{13}{\sqrt{\qmin}}\eps_\M\;,
\end{eqnarray*}
where the last line holds for $\eps_\M\leq1/2$. 
Since $\|\hG-\bG\|_2\leq \eps_\G + (13/\sqrt{\qmin})\eps_\M$, 
we show that $\hV^{\tH}$ and $\bV$ are close using the perturbation analysis 
for robust power method from \cite{AGH12}. 
\begin{theorem}[Restatement of Theorem 5.1 by \cite{AGH12}]
	\label{thm:aghkt}
  Let $G=\sum_{i \in [r]} \lambda_i (v_i \otimes v_i \otimes v_i)+E$, where $\|E\|_2\leq C_1\frac{\lambda_{min}}{r}$. Then the tensor power-method after $N\geq C_2 (\log r+ \log \log \left(\frac{\lambda_{max}}{\|E\|_2}\right)$, generates vectors $\hv_i, 1\leq i\leq r$, and $\hlambda_i, 1\leq i\leq r$, s.t., 
  \begin{equation}
    \label{eq:thm_tnsr}
    \|v_i-\hv_{P(i)}\|_2\leq 8\|E\|_2/\lambda_{P(i)},\quad |\lambda_{i}-\hlambda_{P(i)}|\leq 5\|E\|_2. 
  \end{equation}
where $P$ is some permutation on $[r]$. 
\end{theorem}
Applying the above theorem to $\hG$ and $\bG$, we get that 
$\|\hV^{\tH}-\bV\|_2 \leq 8 \sqrt{\r\,\qmax} (\eps_\G+(13/\sqrt{\qmin})\eps_\M)$. 
Notice that to apply the perturbation analysis, 
it is crucial that we use the fact that  $\bG$ has an orthogonal decomposition. 
Similarly, we can show that 
\begin{eqnarray*}
	\Big|\hlambda_i - \frac{1}{\sqrt{\q_i}}\Big| &\leq& 5\eps_\G + \frac{65}{\sqrt{\qmin}}\,\eps_\M\;.
\end{eqnarray*}

\subsection{Proof of Lemma \ref{lem:M2sample}}

Let $E=\cP_{\Omega_2}(S_2 - \E[S_2]) = S_2 - \E[S_2] - \diag(S_2 - \E[S_2])$, 
 and we bound each term separately using concentration inequalities. 
Define the random matrix 
$E^{(1)} \equiv S_2-\E[S_2] = \frac{2}{|\cS|} \sum_{t\in\{1,\ldots,|\cS|/2\}} \Big(x_tx_t^T - \E[x_tx_t^T] \Big)$. 
We apply the following matrix Bernstein bound for sum of independent sub-exponential random matrices 
with $X_t=x_tx_t^T-\E[x_tx_t^T]$.  
\begin{thm}[Theorem 6.2 of \cite{Jo11}]
	Consider a finite sequence $\{X_t\}$ of independent, random, self-adjoint matrices with dimension $\n$. 
	Assume that 
	$$ \E[X_t]=0 \;\;\;\text{ and }\;\;\; \E[X_t^k] \,\preceq\, \frac{k!}{2} R^{k-2} A_t^2 \;\;\text{ for }\,k=2,3,4,\ldots $$
	Compute the variance parameter $\sigma^2 \equiv \|\sum_{t} A_t^2 \|_2$. 
	Then for all $ a\geq0$, 
	\begin{eqnarray*}
		\prob \Big( \big\| \sum_t X_t \big\|_2 \geq a \Big) &\leq& \n \, \exp \Big( \frac{- a^2/2}{\sigma^2 + Ra } \Big) \;. 
	\end{eqnarray*}
	\label{thm:matrixbernstein}
\end{thm}
The random matrix we defined $X_t$ is zero-mean, and satisfies 
$\E[X_t^k] \preceq \E[(x_tx_t^T)^k] = \ell^{k-1} \E[x_tx_t^T] $, 
which follows form the fact that $x_t^Tx_t=\ell$ almost surely.  
We can prove the inequality $\E[X_t^k] \preceq \E[(x_tx_t^T)^k]$ 
via induction. Let $\bX\equiv \E[x_tx_t^T]$. 
When $k=1$, $\E[X_t] = 0  \preceq \E[(x_tx_t^T)]$, since   
$\bX$ is a convex combination of positive semidefinite matrices. 
When $k=2$, $\E[X_t^2] = \E[(x_tx_t^T)^2] -\bX^2 \preceq \E[(x_tx_t^T)^2]$, 
since $\bX$ is positive semidefinite. 
Suppose $0\preceq \E[(x_tx_t^T)^k - X_t^k] $ for some $k\geq1$, 
then this implies 
$\E[X_t\big((x_tx_t^T)^k - X_t^k\big)X_t] \succeq 0$. 
It follows that 
$$\E\big[\,(x_tx_t^T)^{k+2} +\bX(x_tx_t^T)^{k}\bX - (x_tx_t^T)^{k+1}\bX- \bX(x_tx_t^T)^{k+1}-X_t^{k+2}\,\big] \succeq 0\;,$$
which implies 
\begin{eqnarray*}
	\E\big[\,(x_tx_t^T)^{k+2} -X_t^{k+2}\,\big] &\succeq&  \E\big[\,  (x_tx_t^T)^{k+1}\bX+ \bX(x_tx_t^T)^{k+1} -\bX(x_tx_t^T)^{k}\bX\,\big] \\
	&\succeq& \ell^{k-1}\bX(2\ell\id - \bX)\bX \\
	&\succeq& 0\;.
\end{eqnarray*}
This follows from the fact that $\|\bX\|\leq \E[\|x_tx_t^T\|]= \ell$. 
By induction, 
this proves the desired claim that $\E[X_t^k] \preceq \E[(x_tx_t^T)^k] = \ell^{k-1} \E[x_tx_t^T] $, for all $k$. 
Then, the condition in Theorem \ref{thm:matrixbernstein} is satisfied with 
$R=\ell$, $A_t^2=\ell\,\E[x_tx_t^T]$, and $\sigma^2 = (|\cS|/2)\, \ell\,  \|\E[x_tx_t^T]\|_2$, 
which gives 
\begin{eqnarray*}
	\prob\Big( \frac{|\cS|}{2}\big\| \, E^{(1)} \, \big\|_2 \geq a\Big) &\leq& \n \exp \Big( \frac{-a^2/2}{|\cS| \ell \|\E[x_tx_t^T]\|_2/2 + \ell a}\Big) \;.
\end{eqnarray*}
When $|\cS| \geq (2\ell \log(\n/\delta)) / \|\E[x_tx_t^T]\|_2$ as per our assumption, 
the first term in the denominator dominates for $a=\sqrt{ 2|\cS| \,\ell \,\|\E[x_tx_t^T ]\|_2\,\log(\n/\delta)}$. 
This implies that with probability at least $1-\delta$, 
\begin{eqnarray*}
	\|E^{(1)}\|_2  &\leq& \sqrt{ \frac{4 \,\ell \,\|\E[x_tx_t^T ]\|_2\,\log(\n/\delta)}{|\cS|}}\;,
\end{eqnarray*}
and since $\|\E[x_tx_t^T]\|_2 = \| \frac{\l(\l-1)}{\n(\n-1)}\cP_{\Omega_2}(\M_2) + (\l/\n)\id \|_2\leq (\l^2/\n^2)\sigma_1(\M_2) +(\ell/\n)$, this gives the desired bound.

Now let $E^{(2)} \equiv \diag(S_2-\E[S_2])$, where each diagonal term is 
distributed as binomial distribution Binom($|\cS|/2,\ell/\n$). 
When $|\cS| \geq (2\n/\ell)\log(|\cS|/\delta)$ standard Bernstein inequality gives 
\begin{eqnarray*}
	\|E^{(2)}\|_2 \;\leq\; \max_{i\in[\n]} E^{(2)}_{ii} &\leq& \sqrt{\frac{4\ell\,\log(\n/\delta)}{\n\,|\cS|} }\;,
\end{eqnarray*}
for all $i\in[\n]$ with probability at least $1-\delta$. 
Together we have the desired upper bound on  $\|E^{(1)}+E^{(2)}\|_2$.

In the case of  $\|E_i\|$, similar concentration of measure shows that 
$|E_{ij}| \leq \sqrt{\frac{8\,\l^2\,\log(\n/\delta)}{\n^2\,|\cS|} }$ with probability at least $1-\delta$ for all $j\neq i$. 
This gives 
\begin{eqnarray*}
	\|\cP_{\Omega_2}(E)_i\| &\leq& \sqrt{\frac{8\,\l^2\,\log(\n/\delta)}{\n\,|\cS|}}\;.
\end{eqnarray*}

%
%
\subsection{Proof of Lemma \ref{lem:M3sample}}
 Let $\hH_{abc} = \frac{n(n-1)(n-2)}{\l(\l-1)(\l-2)}\frac{2}{|\cS|}\sum_{t=1+|\cS|/2}^{|\cS|} Y^t_{abc}$ 
 where $Y^t_{abc}=\sum_{(i,j,k)\in\Omega_3}x_{t,i}x_{t,j}x_{t,k}\hQ_{ia}\hQ_{jb}\hQ_{kc}$, 
 and $\hQ=\tU_{\M_2}\tSigma_{\M_2}^{-1/2}$. Then, 
 \begin{eqnarray*}
 	Y^t_{abc} &=& \<x_t,\hQ_a\>\<x_t,\hQ_b\>\<x_t,\hQ_c\> - \<x_t,\hQ_a\>\sum_{i\in[\n]}(x_{t,i}^2\hQ_{ib}\hQ_{ic})
		- \<x_t,\hQ_b\>\sum_i(x_{t,i}^2\hQ_{ia}\hQ_{ic}) \\
		&&\;\;\;\; - \<x_t,\hQ_c\>\sum_i(x_{t,i}^2\hQ_{ia}\hQ_{ib}) 
		+ 2 \sum_{i\in[\n]} x_{t,i}^3 \hQ_{ia} \hQ_{ib} \hQ_{ic}\;.
 \end{eqnarray*}
 We claim that $|Y^t_{abc}|\leq 6 \l^3\mu_1^3 \r^{3/2} \n^{-3/2} (1-\eps_\M)^{-3/2}\sigma_r(\M_2)^{-3/2}$. 
 Since $x_t$ has only $\ell$ non-zero entries and by incoherence of $\mu(\hM_2)=\mu_1$, 
 we get that $|\<x_t,\hQ_a\>| \leq \l \mu_1 \sqrt{\r/(\n(1-\eps_\M)\sigma_r(\M_2))}$. 
Similarly, $|\sum_{i\in[\n]}(x_{t,i}^2\hQ_{ib}\hQ_{ic})| \leq \l \mu_1^2 (\r/\n) (1-\eps_\M)^{-1}\sigma_r(\M_2)^{-1}$ and 
 $|\sum_{i\in[\n]}(x_{t,i}^3\hQ_{ia}\hQ_{ib}\hQ_{ic})| \leq \l \mu_1^3 (\r/\n)^{3/2}(1-\eps_\M)^{-3/2}\sigma_r(\M_2)^{-3/2}$. 
 
 Applying Hoeffding's inequality to $\hH_{abc}$, we get that 
 \begin{eqnarray*}
 	\big| \hH_{abc}-H_{abc}\big|&\leq& \frac{48\,\r^{3/2}\,\mu_1^3\,\n^{3/2}}{\sigma_r(\M_2)^{3/2}}\sqrt{\frac{\log(2/\delta)}{|\cS|}}\;,
 \end{eqnarray*}
 for $\eps_\M\leq1/2$ with probability at least $1-\delta$, 
 where $H_{abc}=\cP_{\Omega_3}(\M_3)[\tU_{\M_2}\tSigma_{\M_2}^{-1/2},[\tU_{\M_2}\tSigma_{\M_2}^{-1/2},[\tU_{\M_2}\tSigma_{\M_2}^{-1/2}]$.
 
%
%
\section{Proof of Theorem \ref{thm:rankcentrality} for the error bound of {\sc RankCentrality}}

The proof builds on key technical Lemma from \cite{NOS12}. Recall that the comparison 
graph $G = ([n], E)$ has $N=|E|$ pairs/edges  and the transition matrix for a random walk on
this graph $G$ for mixture component $a, 1\leq a \leq r$ is $\tp^{(a)} = [\tp_{i,j}^{(a)}] \in [0,1]^{n\times n}$
which depends on $\tP_a$, the estimation of $P_a$ obtained by the algorithm in phase 1. 

The Markov chain $\tp^{(a)}$ is designed in such a way that if $\tP_a$ were indeed exactly equal 
to $P_a$, based on the true model parameters $\bw^{(a)}$, then the following holds (easy to check): 
(a) Markov chain $\tp^{(a)}$ is irreducible and aperiodic as long as $G$ is connected, (b) the 
stationary distribution $\tpi^{(a)}$ is equal (proportional) to $\bw^{(a)}$ (without loss of generality, we
assume that $\bw^{(a)}$ are such that they sum up to $1$). The above fact primarily holds because
in this ideal scenario the correspond Markov chain is a reversible Markov chain with the desired stationary
distribution. In reality, the Markov chain $\tp^{(a)}$ is an approximation of the ideal scenario and we need
to quantity the error due to estimation error of $\tP_a$. This is precisely what we shall do next using 
the following Lemma of \cite{NOS12}. 

\begin{lemma}[Lemma 2 in \cite{NOS12}]
	For a Markov chain $\tp$ and 
	an aperiodic, irreducible and  reversible Markov chain $p$ with the stationary distribution $\pi$, let $\Delta=\tp-p$ 
	and let $\tpi^{(t)}$ be the distribution at time $t$ according to the Markov chain $\tp$ when started with initial distribution $\tpi^{(0)}$. 
	Then, 
	\begin{eqnarray*}
		\frac{\|\tpi^{(t)}-\pi\|}{\|\pi\|} &\leq&  \rho^t \frac{\|\tpi^{(0)}-\pi\|}{\|\pi\|} \sqrt{\frac{\pi_{\rm max}}{\pi_{\rm min}}} + \frac{1}{1-\rho}\|\Delta\|_2 \sqrt{\frac{\pi_{\rm max}}{\pi_{\rm min}}} \;,
	\end{eqnarray*} 
	where $\pi_{\rm min}=\min_{i\in[n]} \pi_i$, $\pi_{\rm max}=\max_{i\in[n]} \pi_i$, $\rho=\lambda_{\rm max}(p)+\|\Delta\|_2 \sqrt{\pi_{\rm max}/\pi_{\rm min}}$, 
	and $\lambda_{\rm max}(p) = \max\{|\lambda_2(p)|,\ldots,|\lambda_n(p)|\}$ is the second largest eigenvalue of $p$ in absolute value. 
\end{lemma}

To prove the desired bound, we need control on the quantities $\|\Delta\|_2$ and $\rho$.  
We know that the spectral norm is bounded by $\|\Delta\|_2\leq \varepsilon$ as per our assumption. 
The following lemma provides a lower bound on $1-\rho$. 
\begin{lemma} 
\label{lem:boundrho}
For $\varepsilon\leq (1/4)\xi \b^{-5/2} (\dmin/\dmax)$, we have 
	\begin{eqnarray*}
		1-\rho &\geq & \frac{1}{4} \frac{\xi}{\b^2} \frac{\dmin}{\dmax} \;.
	\end{eqnarray*}
\end{lemma}
Note that we defined $\b\equiv \pi_{\rm max}/\pi_{\rm min}$ and also it follows from the definition that 
$\|\pi\| \geq 1/\sqrt{n}$ and $\|\tpi^{(0)}-\pi\|\leq2$.
Substituting the bounds on $\|\Delta\|_2\leq \varepsilon$ and $1-\rho$, we get that 
there exists positive numerical constant $C,C'$ such that 
for $t\geq C' \log\big(n/\varepsilon\big) /\log(\rho)$, 
\begin{eqnarray*}
	\frac{\|\tpi^{(t)}-\pi\|}{\|\pi\|} &\leq& \frac{C\,\b^{5/2}\,\dmax}{\xi\, \dmin}\, \varepsilon \;.
\end{eqnarray*}
The necessary number of iterations can be further bounded by 
$t\geq C'(\b^2\dmax/(\xi\dmin))\big(\log(1/\varepsilon) + \log(n)\big)$ using Lemma \ref{lem:boundrho}.
This proves that the bound in Theorem \ref{thm:rankcentrality}, 
for $\varepsilon \leq (1/4)\xi \b^{-5/2}(\dmin/\dmax)$.

Now we are left to prove Lemma \ref{lem:boundrho}. 
From the definition of $\rho$, 
\begin{eqnarray*}
	1-\rho &\geq& 1-\lambda_{\rm max}(p) - \varepsilon\sqrt{\b}\;. 
\end{eqnarray*}
We will show that 
\begin{eqnarray} 
	1-\lambda_{\rm max}(p) &\geq& \frac{\xi\dmin}{2\b^2\dmax} \;,
	\label{eq:rhobound}
\end{eqnarray}	
 where the desired bound follows for 
$\varepsilon \leq (1/4)\xi\dmin/(\b^{5/2}\dmax) $ as per our assumption.
To prove \eqref{eq:rhobound}, we use the comparison theorems (cf. see \cite{NOS12}), which bound the 
spectral gap of the Markov chain $p$ of interest, by comparing it to a more tractable Markov chain. 
We use the simple random walk on the undirected graph $G([n],E)$ as a reference. 
Define the transition matrix of a simple random walk as 
\begin{eqnarray*}
	Q_{ij} &=& \left\{ \begin{array}{rl}
	\frac{1}{d_i}& \text{ for }(i,j)\in E\;,\\
	0& \text{ otherwise}\;,
	\end{array}
	\right.
\end{eqnarray*}
where $d_i$ is the degree of node $i$. 
The stationary distribution of this Markov chain is $\mu_i = d_i/\sum_j d_j$, 
and $Q$ is reversible since the detailed balance equation is satisfied, i.e. $\mu_iQ_{ij} = 1/(\sum_jd_j) = \mu_j Q_{ji}$ for all $(i,j)\in E$. 
The following key lemma provides a bound on the spectral gap of $p$ with respect to the spectral gap of $Q$. 
\begin{lemma}[Lemma 6 in \cite{NOS12}]
	Let $Q,\mu$ and $p,\pi$ be reversible Markov chains on a finite set $[n]$ representing random walks on a graph $G([n], E)$, 
	i.e. $p_{ij}=0$ and $Q_{ij}=0$ if $(i,j)\notin E$. For 
	$\alpha \equiv \min_{(i,j)\in E} \{\pi_ip_{ij}/(\mu_iQ_{ij})\} $ and $\beta\equiv \max_{i\in\cV}\{\pi_i/\mu_i\}$, 
	\begin{eqnarray*}
		\frac{1-\lambda_{\rm max}(p)}{1-\lambda_{\rm max}(Q)} &\geq &\frac{\alpha}{\beta}\;.
	\end{eqnarray*}
\end{lemma}
We have defined $\xi\equiv1-\lambda_{\rm max}(Q)$, and $\alpha$ and $\beta$ can be bounded as follows. 
\begin{eqnarray*}
	\alpha &=& \min_{(i,j)\in E}  \frac{\pi_ip_{ij}}{\mu_iQ_{ij}} \\
		&\geq& \min_{(i,j)\in E} \frac{w_iw_j\sum_{k\in [n]}d_k}{\dmax (w_i+w_j)\sum_{k\in\cV}w_k}\;, \text{ and }\\
	\beta &=& \max_{i\in [n]} \frac{\pi_i}{\mu_i}	\\
		&\leq& \max_{i\in [n]} \frac{w_i\sum_k d_k}{d_i\sum_k w_k}\;.
\end{eqnarray*}
Hence, $\alpha/\beta \geq \dmin/(2\dmax \b^2)$. 
This proves the desired bound in \eqref{eq:rhobound}. 

%
%
\section{Proof of Remark \ref{rem:example}}
Assuming $w^{(a)}_i$'s are drawn uniformly at random from the interval $[1,2]$  
and $G(\cV,\cE)$ drawn from the Erd\"os-R\'enyi model with average degree $\bd\geq \log n$, 
we want to bound the singular values and the incoherence of $\M_2=PQP^T$. 
Define a matrix $\tilde{M}=Q^{1/2}P^TPQ^{1/2}\in\reals^{2\times 2}$. 
Since $\tilde{M}$ and $\M_2$ have the same set of non-zero singular values, 
we analyze the spectrum of $\tilde{M}$. 

For our example, $\tilde{M}=\frac12 P^TP$. 
Define $P_1$ and $P_2$ be the two columns of $P$ such that $P=[P_1\, P_2]$, 
then using McDiarmid's inequality we get that, conditioned on the graph $G$ with $\n$ edges and maximum degree $\dmax$,   
\begin{eqnarray}
  \big|\,P_1^TP_2  \,\big|   &\leq & \varepsilon \n \;,\nonumber\\
  \big|\, \|P_1\|^2- (\ln(3486784401/68719476736)+3)\n \,\big| &\leq & \varepsilon \n \;,\nonumber\\
  \big|\, \|P_2\|^2- (\ln(3486784401/68719476736)+3)\n \,\big| &\leq & \varepsilon \n \;. \label{eq:exconcentration}
\end{eqnarray}
with high probability for any positive constant $\varepsilon>0$. 
We provide a proof for $\|P_1\|^2$, and the others follow similarly. 
Conditioned on the graph $G$, $\|P_1\|^2=\sum_{(i,j)\in\cE} (\,\big(w_j^{(1)}-w_i^{(1)})/(w_i^{(1)}+w_j^{(1)})\,\big)^2=f(w^{(1)}_1,\ldots,w^{(1)}_\n)$ 
is a function with bounded difference:
$$ \sup_{w^{(1)}_1\ldots,w^{(1)}_n,v^{(1)}_i} \big|\,f(w^{(1)}_1,\ldots,w^{(1)}_{i-1},w^{(1)}_{i},w^{(1)}_{i+1},\ldots,w^{(1)}_n) - 
f(w^{(1)}_1,\ldots,w^{(1)}_{i-1},v^{(1)}_{i},w^{(1)}_{i+1},\ldots,w^{(1)}_n) \,\big|\leq \dmax \;.$$
It follows that 
\begin{eqnarray*}
      \prob\Big( \big|\, \|P_1\|^2 - \E[\|P_1\|^2] \,\big|\geq \varepsilon \n \,\Big|\, G\Big) &\leq& 2\exp \Big\{-\frac{2\varepsilon^2\n^2}{\dmax^2 n}\Big\}\;.
\end{eqnarray*}
For Erd\"os-R\'enyi random graphs with $\bd\geq\log n$,  we know that 
$\dmax=\Theta(\bd)$ and $N=\Theta(\bd \n)$ with high probability. 
Also, it is not too difficult to compute $\E[\|\P_1\|^2] = \ln(3486784401/68719476736)+3 \simeq 0.0189$.
It follows that with high probability, \eqref{eq:exconcentration} holds. 

Given \eqref{eq:exconcentration}, we can decompose the matrix as 
\begin{eqnarray*}
      \tilde{M} = \begin{bmatrix}(\ln(3486784401/68719476736)+3)\n& 0 \\ 0& (\ln(3486784401/68719476736)+3)\n\end{bmatrix} + \Delta\;,
\end{eqnarray*}
where $\|\Delta\|_2\leq 2\varepsilon$. 
It follows that 
$\sigma_1(\tilde{M})\leq 0.02 \n$ and 
$\sigma_2(\tilde{M})\geq 0.017 \n$. 
Choosing $\varepsilon=0.001 \n$, this proves the desired bound. 

To bound the incoherence, consider the SVD of $\M_2=USU^T = PQP^T$. 
There exists a orthogonal matrix $R$ such that $US^{1/2}=PQ^{1/2}R$. 
Then, the $i$-th row of $U$ is $U_i=e_i^TPQ^{1/2}RS^{-1/2}$. 
We know, $Q=diag(1/2,1/2)$, $S=\diag(\sigma_1(\M_2),\sigma_2(\M_2))$, and $RR^T=R^TR=\id$.  
It follows that 
\begin{eqnarray*}
  \mu(\M_2) &=& \max_i \sqrt{\n/2} \|U_i\| \\
    &\leq& \max_i \sqrt{\n/2} \sqrt{P_{i1}^2+P_{i2}^2}(1/\sqrt{2})\sqrt{1/\sigma_2(\M_2)}  \\
  &\leq& 15 \;.  
\end{eqnarray*}

\end{document}